\documentclass[11pt, a4paper]{article}

\usepackage[margin=1in]{geometry}
\usepackage{amsmath}
\usepackage{amssymb}
\usepackage{amsthm}
\usepackage{graphicx}
\usepackage{booktabs}
\usepackage[numbers]{natbib}
\usepackage{url}
\usepackage[hidelinks]{hyperref}

\graphicspath{{../docs/figures/}{figures/}}

\newtheorem{proposition}{Proposition}
\newtheorem{lemma}[proposition]{Lemma}
\newtheorem{theorem}[proposition]{Theorem}
\newtheorem{corollary}[proposition]{Corollary}
\theoremstyle{remark}
\newtheorem{remark}[proposition]{Remark}
\theoremstyle{plain}

\newcommand{\Abar}{\bar{A}}
\newcommand{\Bbar}{\bar{B}}

\title{A Log-Timescale Perturbation Bound for Selective\\ State Space Models at Every Sequence Length}

\author{
  \begin{tabular}{c@{\hspace{4em}}c}
    Jack Humphries & Daniel Midgley \\
    \small Department of Computer Science & \small Department of Mathematics \\
    \small University of Surrey & \small University of Surrey \\
    \small\texttt{jh02956@surrey.ac.uk} & \small\texttt{dm01602@surrey.ac.uk}
  \end{tabular}
}

\date{}

\begin{document}
\maketitle

\begin{abstract}
\noindent Mamba models deployed on streaming data, such as limit order book updates, run far past
the sequence lengths they were trained on. A guarantee on how much a small change of input can
move the hidden state therefore has to hold at every length. The closest existing bound we know of
measures the change of the timescale additively, and it grows linearly with the sequence length
unless the timescale is bounded away from zero. In this paper, we measure the change on a log
scale instead. The key observation is that the decay factor of each step then moves by at most a
fixed fraction of that step's input weight, so the per-step errors cannot pile up. Using this, we
prove that two runs of a selective-scan cell from a common start stay within a bound that holds
at every length, shrinks to zero with the size of the change and needs no lower bound on the
timescale. We give a construction showing that, for small changes, this bound is within a factor
of about $2$ of the best possible, and we extend it to the update used in Mamba's released code
when the timescale is bounded above. We then evaluate block-level bounds built from the theorem
on two small models trained on one day of order book data. With a bounded timescale, a bound
built on the existing analysis is tighter than ours for at most the first $8$ steps. With Mamba's
softplus timescale, whose certified lower bound is as small as $10^{-21}$, it is looser at every
length. Our bounds hold for every input to a block, but they remain more than five orders of
magnitude above the changes that gradient attacks find.
\end{abstract}

\section{Introduction}
\label{sec:intro}

Limit order book updates arrive as an unbounded stream. A selective state space model such as
Mamba \citep{gu2023mamba} can carry its state along such a stream far beyond its training windows.
A bound on how far a small change of input moves that state then has to hold at every length. We
prove such a bound for one cell, a scalar recurrence inside the selective scan, and measure how
tight it is.

The closest bound we know of, Cao et al.'s Theorem~3.3(ii) \citep{cao2026control}, grows linearly
with the position. The growth comes from its time-warp term, which accounts for the change of the
timescale, the step size that the model computes from each input. Their Remark~3.4 caps the growth
under a floor, a positive lower bound $\Delta_{\mathrm{lo}}$ on the timescale, but the cap scales
like $1/\Delta_{\mathrm{lo}}$ as the floor shrinks. Mamba's softplus timescale has a floor only
when its pre-activations are bounded below, and on one of our trained models we certify a floor of
about $10^{-21}$.

We measure the change of the timescale on a log scale, whereas Cao et al.\ measure it additively.
In their bound, both the linear growth and the need for a floor come from the additive measure. We
require the timescales $\Delta_t$ and $\tilde\Delta_t$ of a nominal and a perturbed run to differ
by at most a factor $e^{\rho}$ at every step, that is,
$|\log\Delta_t - \log\tilde\Delta_t| \le \rho$. We call this the log condition.

With an exact zero-order hold, which our models use, each step of the cell replaces the state by a
weighted average of itself and an input-dependent equilibrium. The weight on the equilibrium,
which we call the input weight, tends to zero with the timescale. Under the log condition, the two
runs' weights on the previous state differ by at most $e^{\rho} - 1$ times the input weight
(Lemma~\ref{lem:ratio}). Under an additive bound, this difference does not shrink with the input
weight and can exceed it at small timescales (Figure~\ref{fig:mechanism}).

With Lemma~\ref{lem:ratio}, the gap between the two runs obeys an inequality of the same form as
the update of the state, with a bounded equilibrium. The standard invariant argument for convex
recursions, which Horbatko states for a Mamba channel \citep[Lemma~4.4]{horbatko2026sessa}, bounds
this gap at every length, as it bounds the state. From a common start, the variation-of-constants
argument of Cao et al.'s Lemma~B.2 also gives this bound after a change of time
variable (Appendix~\ref{app:cao}).

Our main result, Theorem~\ref{thm:cell}, assumes the log condition, a bound $\eta$ on the change
of the equilibrium and a bound $M$ on the equilibrium of the perturbed run. Two runs that start
from the same state in $[-M, M]$ then stay within $\eta + 2M(e^{\rho} - 1)$ of each other at every
step, with no floor on the timescale. For every $M$, $\eta$ and $\rho$,
Proposition~\ref{prop:lower} constructs such runs whose gaps come arbitrarily close to
$\eta + 2M\tanh(\rho/2)$. From a common start, the theorem's constant is therefore within a factor
$e^{\rho} + 1$ of the best possible, a factor that tends to $2$ as $\rho \to 0$.

If $\rho$ is derived from Cao et al.'s additive bound and floor, our time-warp term
$2M(e^{\rho} - 1)$ is no larger in each cell than that of their Remark~3.4. It is also no smaller
than the limit, as the length grows, of a slightly sharper form of their bound, and the three are
close when the floor is small (Section~\ref{sec:compare} and Appendix~\ref{app:cao}). The gain in our experiments comes instead from bounding
$\log\Delta$ directly from the bound on the timescale's pre-activation.

A gradient search in one cell, with both runs started from $0$, finds no gap above the value that
the construction of Proposition~\ref{prop:lower} approaches (Section~\ref{sec:one-cell}). From
random initial sequences it comes within $0.4$ per cent of that value at the longest lengths
searched, and closer still from initial sequences shaped like the construction.

We then train a small two-block model on one day of limit order book data, once with a timescale
bounded to a fixed range and once with Mamba's softplus timescale. From the theorem we build a
bound for each block on how far its output can move when its normalised input changes
(Section~\ref{sec:certificate}). With the bounded timescale, the same bound with an additive
time-warp term of Cao et al.'s form in place of ours is tighter for at most the first $8$ steps.
Without a floor, it then grows linearly with the length. Under the floor that we certify, it
levels off at a multiple of ours that is set mainly by the chosen timescale range
(Section~\ref{sec:floor-exp}). With the softplus timescale, it is looser than ours at every
length, and the certified floor is too small to stop its linear growth at any length we test.

Gradient attacks give lower bounds on the true sensitivity, and our block bounds, which hold for
every input of a block, lie far above them. Along a given input the realised timescale is known,
and the log and additive bounds on its change agree to first order. The log scale therefore
matters for bounds that hold for every input, where no realised timescale is available
(Section~\ref{sec:sens-experiments}). Code and the recorded runs are at
\url{https://github.com/jackdoesjava/mamba-v-lob}.

Section~\ref{sec:setting} sets up the cell. Section~\ref{sec:sensitivity} proves the theorem,
states the lower bound and extends the theorem to the exponential-Euler step of the released Mamba
implementations, with the proofs of the lower bound and the extension in Appendix~\ref{app:proofs}.
Section~\ref{sec:certificate} builds the block bounds, and Section~\ref{sec:experiments} reports
the measurements.

\section{Setting}
\label{sec:setting}

\subsection{The selective cell}
\label{sec:block}

A selective state space model computes a timescale $\Delta_t > 0$ and matrices $B_t$ and $C_t$
from the layer input at every position $t$. It uses the timescale to discretise a fixed diagonal
matrix $A$ with negative diagonal entries \citep{gu2023mamba}. We certify a Mamba block that uses
the exact zero-order hold for this discretisation. The block applies a layer normalisation, an
input projection, a causal depthwise convolution with a SiLU, the projections that produce
$\Delta$, $B$ and $C$, and the scan. A SiLU gate, an output projection and a residual connection
follow, and Appendix~\ref{app:block} gives the block in full.

Because $A$ is diagonal, the scan splits into independent scalar recurrences, one for each channel
$i$ and state index $n$. We call each of them a cell. A cell updates a scalar state $h_t$ from the
input $u_t$ of its channel, and we write $\lambda = |A_{i,n}| > 0$ for its decay rate. We state
our results on the scan for one cell.

We consider two maps from the timescale pre-activation $z$ to $\Delta$. The bounded map
$\Delta = \Delta_{\min}(\Delta_{\max}/\Delta_{\min})^{\operatorname{sigmoid}(z)}$ keeps $\Delta$
in $[\Delta_{\min}, \Delta_{\max}]$. Mamba's map $\Delta = \mathrm{softplus}(z)$ is positive, but
it has no positive lower bound over real $z$. We call a positive lower bound on the timescale a
floor.

Under the exact zero-order hold, one step of a cell uses
\begin{equation}
\label{eq:zoh}
  \Abar_t = e^{-\lambda\Delta_t},
  \qquad
  \Bbar_t = \frac{e^{-\lambda\Delta_t} - 1}{-\lambda}\, B_t = (1 - \Abar_t)\,\frac{B_t}{\lambda},
\end{equation}
and with $c_t = B_t u_t/\lambda$ the update $h_t = \Abar_t h_{t-1} + \Bbar_t u_t$ becomes
\begin{equation}
\label{eq:convex}
  h_t = \Abar_t\, h_{t-1} + (1 - \Abar_t)\, c_t .
\end{equation}
Since $\Abar_t \in (0,1)$ for every input, each step is a convex combination of the previous state
and $c_t$. We call $c_t$ the equilibrium and $1 - \Abar_t$ the input weight of step $t$. Gu and Dao
\citep[Theorem~1]{gu2023mamba} give~\eqref{eq:convex} as a gated recurrence for $\lambda = 1$ and
$B_t = 1$ with a softplus timescale.

The released implementations of Mamba use a different input term, which we treat in
Section~\ref{sec:euler}.

\subsection{The forward invariant}

Because each step is a convex combination, a bound on the equilibria and on the initial state
bounds the state at every step.

\begin{proposition}[Forward invariant]
\label{prop:invariant}
If $|c_t| \le M$ for every $t \ge 0$, then $|h_t| \le \max(|h_{-1}|, M)$ for every $t \ge 0$.
\end{proposition}

\begin{proof}
We argue by induction on $t$: if $|h_{t-1}| \le \max(|h_{-1}|, M)$, then
$|h_t| \le \Abar_t |h_{t-1}| + (1 - \Abar_t)|c_t| \le \max(|h_{-1}|, M)$.
\end{proof}

Proposition~\ref{prop:invariant} is Lemma~4.4 of Horbatko \citep{horbatko2026sessa}, stated there
for a diagonal Mamba channel. For $h_{-1} = 0$ it also follows from the state bound of Cao et al.'s
Theorem~3.3(i) and Lemma~B.1 \citep{cao2026control}. The reading of a convex gated update as a
leaky integrator with a learned step goes back to Tallec and Ollivier \citep{tallec2018warp} and
Gu et al.\ \citep{gu2021lssl}.

Since $B_t$ and $u_t$ are activations, the hypothesis $|c_t| \le M$ has to hold for every input the
model can receive. Within the scan and the readout $y_t = C_t h_t + D u_t$ that follows it, only
$A$ and the skip weight $D$ are fixed parameters. The timescale $\Delta_t$, the matrices $B_t$ and
$C_t$ and the scan input $u_t$ are all computed from the block input.

Each block opens with a layer normalisation. By the Laguerre--Samuelson inequality
\citep{laguerre1880methode,samuelson1968deviant}, the output of the normalisation lies in a bounded
set fixed by its own parameters (Lemma~\ref{lem:ln} in Appendix~\ref{app:separate})\@. Interval
arithmetic carries this set through the projections, the convolution and the activations. For the
projection that follows the normalisation, we use the fused bound of Proposition~\ref{prop:fused}
in Appendix~\ref{app:separate},
which follows from the geometry of layer normalisation that Riechers \citep{riechers2024layernorm}
describes. The result is a box, with one interval per coordinate, for the output of each of
these stages, including $B$, $u$ and $z$. We call these the certified boxes. The boxes of $B$ and
$u$ give the radius $M = \sup|Bu|/\lambda$ of each cell. The box $Z$ of the timescale
pre-activation of a channel gives its certified floor $\Delta(\inf Z)$ and certified ceiling
$\Delta(\sup Z)$, the smallest and largest timescales that the box allows.

We have machine-checked~\eqref{eq:zoh}, Proposition~\ref{prop:invariant}, Lemma~\ref{lem:ln} and
the first part of Proposition~\ref{prop:fused} in Isabelle/HOL \citep{nipkow2002isabelle}, and the
theory is in our repository. Lemma~\ref{lem:ratio}, Theorem~\ref{thm:cell} and
Proposition~\ref{prop:lower} are not machine-checked. We compute $M$ for a trained model by
interval propagation in Python code, which we test but have not formally verified.

\section{A sensitivity bound at every length}
\label{sec:sensitivity}

We bound how far apart two runs of one cell can be at any length. The bound holds when their
equilibria are bounded and their timescales differ by at most a factor $e^{\rho}$ at every step.
The nominal run has timescales $\Delta_t > 0$, equilibria $c_t$, transitions
$\Abar_t = e^{-\lambda\Delta_t}$ and states $h_t$ given by~\eqref{eq:convex}. The perturbed run
has $\tilde\Delta_t > 0$, $\tilde c_t$, $\tilde{\Abar}_t = e^{-\lambda\tilde\Delta_t}$ and
$\tilde h_t$. We write
$e_t = h_t - \tilde h_t$ for the error and call $|e_t|$ the gap. The hypothesis on the timescales
is $|\log\Delta_t - \log\tilde\Delta_t| \le \rho$ for every $t$, where $\rho \ge 0$ and $\log$ is
the natural logarithm. We write $\eta = \sup_t |c_t - \tilde c_t|$ for the change of the
equilibrium.

\subsection{The ratio lemma}

Lemma~\ref{lem:ratio} shows that a relative change of the timescale moves the transition by at most
a fixed fraction of the input weight.

\begin{lemma}[Relative perturbation of the transition]
\label{lem:ratio}
For $x > 0$, $s \in \mathbb{R}$ and $\tilde x = x e^{s}$ we have
$|e^{-x} - e^{-\tilde x}| \le |e^{s} - 1|(1 - e^{-x})$. In particular the left-hand side is at
most $(e^{\rho} - 1)(1 - e^{-x})$ whenever $|s| \le \rho$.
\end{lemma}

\begin{proof}
We compare chord slopes of the concave function $y \mapsto 1 - e^{-y}$, which vanishes at $0$
and whose chord slope from the origin, $m(y) = (1 - e^{-y})/y$, is therefore decreasing on
$(0, \infty)$. For $s \ge 0$ we have $\tilde x \ge x$, and the decrease of $m$ gives
$0 \le e^{-x} - e^{-\tilde x} = \tilde x\, m(\tilde x) - x\, m(x) \le (\tilde x - x)\, m(x)
= (e^{s} - 1)(1 - e^{-x})$.

For $s < 0$ we have $\tilde x < x$ and $m(\tilde x) \ge m(x)$, hence
$0 \le e^{-\tilde x} - e^{-x} = x\, m(x) - \tilde x\, m(\tilde x) \le (x - \tilde x)\, m(x)
= (1 - e^{s})(1 - e^{-x})$, which together with $|e^{s} - 1| \le e^{|s|} - 1$ proves both
bounds.
\end{proof}

With $x = \lambda\Delta_t$ and $s = \log(\tilde\Delta_t/\Delta_t)$, Lemma~\ref{lem:ratio} gives
\begin{equation}
\label{eq:abar-ratio}
  \bigl|\Abar_t - \tilde{\Abar}_t\bigr| \le \bigl(e^{\rho} - 1\bigr)\bigl(1 - \Abar_t\bigr),
\end{equation}
and with the runs exchanged it gives the same bound with $1 - \tilde{\Abar}_t$ on the right.
Cao et al.'s Lemma~B.3 bounds the same change by $\lambda|\Delta_t - \tilde\Delta_t|$, which lacks
the factor $1 - \Abar_t$. Under an additive bound on $|\Delta_t - \tilde\Delta_t|$, the term that
each step adds to a bound on the gap does not shrink with the input weight. The later
transitions damp these terms, but their sum grows like $t$ unless a floor makes the damping
geometric.

\begin{figure}[htbp]
  \centering
  \includegraphics[width=\textwidth]{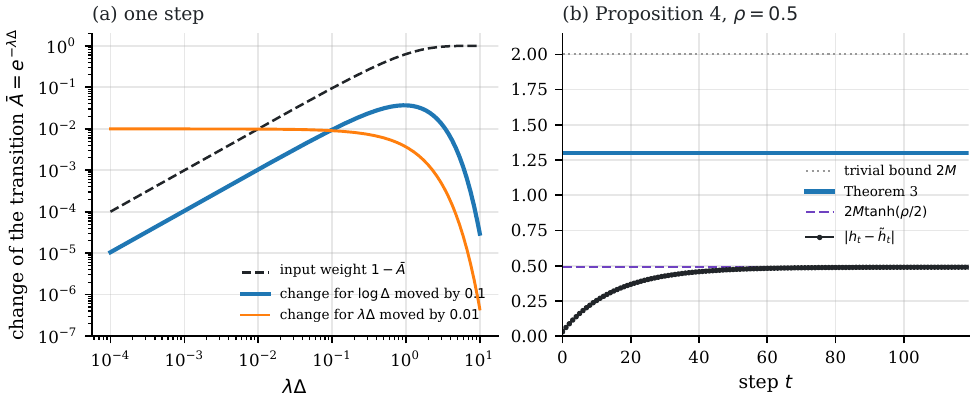}
  \caption{(a) Change of the transition $e^{-\lambda\Delta}$ against $\lambda\Delta$ when
    $\log\Delta$ changes by $0.1$ and when $\lambda\Delta$ changes by $0.01$, with the input weight
    $1 - \Abar$. The two changes of $\lambda\Delta$ are equal at $\lambda\Delta \approx 0.095$, where
    the two curves of the change of the transition cross. The first change stays below $e^{0.1} - 1 \approx 0.105$ times the input
    weight. The second exceeds the whole input weight for $\lambda\Delta$ below about $0.0099$. At
    $\lambda\Delta = 10^{-4}$ it is $99.5$ times the input weight. (b) The gap
    $|h_t - \tilde h_t|$ in the construction of Proposition~\ref{prop:lower} for $\rho = 0.5$,
    $M = 1$ and $\eta = 0$. The nominal $\lambda\Delta_t$ alternates between $0.05\,e^{0.5}$ and
    $0.05$. The gap settles at about $0.4895$, just below the value $2M\tanh(\rho/2) \approx 0.4898$ that
    it approaches as the timescales tend to $0$. The panel shows the bound $E \approx 1.297$
    of Theorem~\ref{thm:cell} and the trivial bound $2M + \eta = 2$.}
  \label{fig:mechanism}
\end{figure}

Figure~\ref{fig:mechanism}(a) shows the two cases at a single step. When $\log\Delta$ changes by a
fixed amount, the change of the transition vanishes with the input weight as $\lambda\Delta \to 0$.
When $\lambda\Delta$ changes by a fixed amount, the change of the transition does not vanish, and
for small $\lambda\Delta$ it exceeds the whole input weight.

\subsection{The cell theorem}

Theorem~\ref{thm:cell} bounds the gap at every step by a constant that depends only on $\eta$, $M$
and $\rho$, or by the initial gap if that is larger.

\begin{theorem}[Cell bound]
\label{thm:cell}
Suppose that $|\log\Delta_t - \log\tilde\Delta_t| \le \rho$ and $|\tilde c_t| \le M$ for every
$t \ge 0$, and that $|\tilde h_{-1}| \le M$. With $\eta = \sup_t |c_t - \tilde c_t|$,
$E = \eta + 2M(e^{\rho} - 1)$ and $P_t = \prod_{k=0}^{t} \Abar_k$, for every $t \ge 0$
\[
  |h_t - \tilde h_t| \le P_t\, |h_{-1} - \tilde h_{-1}| + (1 - P_t)\, E
    \le \max\bigl( |h_{-1} - \tilde h_{-1}|,\ E \bigr).
\]
\end{theorem}

\begin{proof}
We write the error as a convex recursion of the form~\eqref{eq:convex} with a bounded
equilibrium and then argue as for Proposition~\ref{prop:invariant}, which also gives
$|\tilde h_t| \le M$ for every $t \ge -1$. Adding and subtracting $\Abar_t\tilde h_{t-1}$ and
$(1 - \Abar_t)\tilde c_t$ in $h_t - \tilde h_t$ gives
\begin{equation}
\label{eq:error}
  e_t = \Abar_t\, e_{t-1} + (1 - \Abar_t)(c_t - \tilde c_t)
        + (\Abar_t - \tilde{\Abar}_t)(\tilde h_{t-1} - \tilde c_t).
\end{equation}
Since $|\tilde h_{t-1} - \tilde c_t| \le 2M$, the bound~\eqref{eq:abar-ratio} gives
$|e_t| \le \Abar_t\, |e_{t-1}| + (1 - \Abar_t)\, E$. Unrolling this inequality gives
$|e_t| \le P_t |e_{-1}| + E \sum_{k=0}^{t} (1 - \Abar_k) \prod_{r=k+1}^{t} \Abar_r$, where the
sum telescopes to $1 - P_t$.
\end{proof}

The passage from the one-step inequality to the bound of Theorem~\ref{thm:cell} is a standard
contraction estimate, sampled at the end of each step. In the accumulated time
$\tau_t = \sum_{s \le t}\Delta_s$, the cell contracts at the constant rate $\lambda$
(Appendix~\ref{app:cao})\@. The estimate compares a system that contracts at a constant rate $\mu$
with the same system plus a disturbance $d$ \citep{lohmiller1998contraction,bullo2026contraction}.
After a time $\sigma$, the distance between their trajectories is at most $e^{-\mu\sigma}$ times
its initial value plus $(1 - e^{-\mu\sigma})\sup|d|/\mu$. With $\mu = \lambda$,
$|d| \le \lambda E$ and $\sigma = \tau_t$, this is the bound of Theorem~\ref{thm:cell}.

Per step, the cell contracts by the factor $\Abar_t = e^{-\lambda\Delta_t}$, and without a floor
the rate $\lambda\Delta_t$ has no positive lower bound. The bound is still uniform in $t$, because
the disturbance at each step carries the same factor $1 - \Abar_t$ as the contraction, so their
ratio stays bounded as $\Delta_t \to 0$. A bound that treats the transition and the input gain
separately loses this cancellation (Appendix~\ref{app:separate})\@.

The proof uses the hypotheses one step at a time. The bounds $\rho$ and $\eta$ may therefore
depend on $t$, with $E$ replaced by $\max_{s \le t}(\eta_s + 2M(e^{\rho_s} - 1))$.

If $|h_{-1}| \le M + \eta$, Proposition~\ref{prop:invariant} applied to both runs gives
$|h_t| \le M + \eta$ and $|\tilde h_t| \le M$, hence the trivial bound $|e_t| \le 2M + \eta$. For
$M > 0$, the trivial bound exceeds $E$ exactly when $\rho < \log 2$.

\begin{proposition}[Lower bound]
\label{prop:lower}
Let $\lambda > 0$, $M > 0$, $\eta \ge 0$ and $\rho > 0$. For every $\varepsilon > 0$ there are
timescales $\Delta_t, \tilde\Delta_t > 0$ with $|\log\Delta_t - \log\tilde\Delta_t| = \rho$ and
equilibria with $|\tilde c_t| = M$ and $c_t - \tilde c_t = \eta$ for every $t \ge 0$ such that
the two runs from $h_{-1} = \tilde h_{-1} = 0$ satisfy
$|h_t - \tilde h_t| > \eta + 2M\tanh(\rho/2) - \varepsilon$ at some step $t$.
\end{proposition}

The runs in the proof of Proposition~\ref{prop:lower} (Appendix~\ref{app:lower}) satisfy the
hypotheses of Theorem~\ref{thm:cell}. The perturbed equilibrium $\tilde c_t$ alternates between
$M$ and $-M$. The perturbed timescale is $e^{-\rho}$ times the nominal one on the steps with
$\tilde c_t = M$ and $e^{\rho}$ times it on the others. The nominal run therefore has the larger
input weight on the steps with $\tilde c_t = M$ and the smaller one on the others. As
$t \to \infty$ and then the timescales tend to $0$, the gap after a step with $\tilde c_t = M$
tends to $\eta + 2M\tanh(\rho/2)$. Figure~\ref{fig:mechanism}(b) shows the gap for $\rho = 0.5$,
$M = 1$ and $\eta = 0$. It settles close to $2M\tanh(\rho/2)$, below the bound $E$ of
Theorem~\ref{thm:cell}.

Proposition~\ref{prop:lower} limits how far Theorem~\ref{thm:cell} can be improved. Every bound on
the gap from a common start that holds under the hypotheses of the theorem and depends only on
$M$, $\eta$ and $\rho$ is at least $\eta + 2M\tanh(\rho/2)$. Since
$\tanh(\rho/2) = (e^{\rho} - 1)/(e^{\rho} + 1)$, the term $2M(e^{\rho} - 1)$ of $E$ is
$e^{\rho} + 1$ times $2M\tanh(\rho/2)$. The bound $E$ is therefore within a factor $e^{\rho} + 1$
of the smallest such bound, and this factor tends to $2$ as $\rho \to 0$.

From a common start in $[-M, M]$, the largest possible gap thus lies between
$\eta + 2M\tanh(\rho/2)$ and $\min(E, 2M + \eta)$. We do not prove that $\eta + 2M\tanh(\rho/2)$ is
the largest possible gap. The numerical search of Section~\ref{sec:one-cell}, from the common start
$0$, finds none above it.

\subsection{Comparison with Cao et al.}
\label{sec:compare}

We replace the bound on $\log\Delta$ by an additive bound
$|\Delta_t - \tilde\Delta_t| \le \eta_\Delta$ and keep the other hypotheses of
Theorem~\ref{thm:cell}. We also assume a lower bound $\Delta_{\mathrm{lo}} \ge 0$ on both
timescales, which is a floor when it is positive. By the mean value theorem, the transitions then
satisfy $|\Abar_t - \tilde{\Abar}_t| \le \lambda e^{-\lambda\Delta_{\mathrm{lo}}}\eta_\Delta$. We
carry the last term of~\eqref{eq:error} forward with a factor of at most
$e^{-\lambda\Delta_{\mathrm{lo}}}$ per step. From a common start in $[-M, M]$, this gives
\begin{equation}
\label{eq:additive}
  |h_{L-1} - \tilde h_{L-1}| \le \eta + 2M\lambda e^{-\lambda\Delta_{\mathrm{lo}}}\eta_\Delta
    \sum_{k=0}^{L-1} e^{-k\lambda\Delta_{\mathrm{lo}}} .
\end{equation}

We call~\eqref{eq:additive} the additive baseline. It has the structure of Cao et al.'s
Theorem~3.3(ii) and Remark~3.4. It keeps the factor $e^{-\lambda\Delta_{\mathrm{lo}}}$ that their
Lemma~B.3 drops, which makes it slightly smaller than their constant. As Cao et al.\ do, we call
the term of a bound that comes from the change of the timescale its time-warp term. In $E$ this
term is $2M(e^{\rho} - 1)$, and in~\eqref{eq:additive} it is the term with the sum.

Without a floor the sum equals $L$, so the additive baseline grows linearly in the length $L$.
With a floor it saturates at $\eta + 2M\lambda\eta_\Delta/(e^{\lambda\Delta_{\mathrm{lo}}} - 1)$ as
$L \to \infty$. Its time-warp term is then the one-step cost
$2M\lambda e^{-\lambda\Delta_{\mathrm{lo}}}\eta_\Delta$ times
$1/(1 - e^{-\lambda\Delta_{\mathrm{lo}}})$. This factor is of order
$1/(\lambda\Delta_{\mathrm{lo}})$ as the floor shrinks.

Theorem~\ref{thm:cell} needs neither a floor nor an additive bound on the timescales, only a bound
on the change of $\log\Delta$. Both timescale maps give one from a bound on the change of the
pre-activation $z$. By Lemma~\ref{lem:param} in Appendix~\ref{app:proofs}, $d\log\Delta/dz$ is at
most $1$ for the softplus map
and at most $\tfrac14\log(\Delta_{\max}/\Delta_{\min})$ for the bounded map. Cao et al.\ derive the
Lipschitz constant of $\Delta$ that they assume from the same bound on $z$.

With a floor, the additive bound alone also bounds the change of $\log\Delta$, by
$\rho = \log(1 + \eta_\Delta/\Delta_{\mathrm{lo}})$. With this $\rho$, our time-warp term lies
between the time-warp term of the additive baseline as $L \to \infty$ and that of Cao et al.'s
Remark~3.4. This follows from $1 - e^{-x} \le x \le e^{x} - 1$ for $x = \lambda\Delta_{\mathrm{lo}}$
(Appendix~\ref{app:cao})\@. In our experiments, the gain over the additive baseline comes
from bounding $\log\Delta$ directly from the bound on $z$, not through an additive bound on
$\Delta$.

For the bounded map, Remark~\ref{rem:cap} compares the two time-warp terms in one cell once the
additive baseline has saturated.

\begin{remark}[The ratio for the bounded map]
\label{rem:cap}
Let $\Delta_{\mathrm{lo}} \ge \Delta_{\min}$ be a floor, $R = \Delta_{\max}/\Delta_{\min}$ and
$x = \lambda\Delta_{\mathrm{lo}}$, let $Z$ be the certified box of the pre-activation of one
channel, and let a bound $\zeta$ on the change of $z$ within $Z$ give
$\eta_\Delta = \zeta\sup_Z|\Delta'|$ and $\rho = \zeta\sup_Z|(\log\Delta)'|$. To first order in
$\zeta$, the time-warp term of~\eqref{eq:additive} as
$L \to \infty$ is
\[
  \frac{\sup_{Z} |\Delta'|}{\Delta_{\mathrm{lo}}\, \sup_{Z} |(\log\Delta)'|} \cdot \frac{x}{e^{x} - 1}
\]
times ours (Appendix~\ref{app:cao})\@. If $Z$ contains $0$, then
$\sup_Z |(\log\Delta)'| = \tfrac14 \log R$ by Lemma~\ref{lem:param}. With
$p = \operatorname{sigmoid}(z)$ we have $\Delta' = \Delta\,(\log\Delta)' = \Delta_{\min} R^{p}
p(1 - p)\log R$, and since $\Delta_{\min}/\Delta_{\mathrm{lo}} \le 1$ and $x/(e^{x} - 1) \le 1$,
the ratio is then at most $4\max_{p \in (0,1)} R^{p} p(1 - p)$, which for
$R = 100$ is $25.8$, attained at $p \approx 0.83$. With the floor
$\Delta_{\mathrm{lo}} = \Delta(\inf Z)$ that the box certifies, this ratio depends on the trained
weights only through $Z$ and $x$.
\end{remark}

\subsection{The exponential-Euler step}
\label{sec:euler}

The released implementations of Mamba keep the transition
$\Abar_t = e^{-\lambda\Delta_t}$ but replace the input term $\Bbar_t u_t$ by $\Delta_t B_t u_t$
\citep[Section~3.1]{lahoti2026mamba3}. This exponential-Euler step combines the exact transition
with Euler's rule for the input integral. It differs from the exponential Euler method of the
literature on exponential integrators, which for this cell gives~\eqref{eq:zoh}.

The update of this step is still a convex combination of the previous state and an equilibrium.
The equilibrium is now enlarged by the factor $\Psi(\lambda\Delta_t)$, where
$\Psi(x) = x/(1 - e^{-x})$ lies in $(1, 1 + x)$ for $x > 0$. This factor grows without bound in
$\Delta_t$. The analogues of Proposition~\ref{prop:invariant} and Theorem~\ref{thm:cell} for this
step therefore need an upper bound $\Delta_{\mathrm{hi}}$ on the timescale, but they still need no
floor. Corollary~\ref{cor:euler} in Appendix~\ref{app:proofs} gives the analogue of the theorem for
a common zero start. With $G = \Psi(\lambda\Delta_{\mathrm{hi}})$, the bound $E$ holds for this
cell with $\eta$ replaced by $G\eta^{b}$, $M$ by $GM$ and the factor $2$ by $3$. Here $\eta^{b}$
bounds the change of $b_t = B_t u_t/\lambda$.

\section{From the cell to a certificate}
\label{sec:certificate}

Corollary~\ref{cor:block} in Appendix~\ref{app:block} turns Theorem~\ref{thm:cell} into a bound on
the change of a whole block's output. The bound holds in real arithmetic for every input of the
block, and we call such a bound a certificate. It needs no input data, so we also call it
data-free.

The corollary follows a perturbation of the normalised input through the stages of the block in
order. It composes Theorem~\ref{thm:cell} with sup-norm Lipschitz constants of the other stages,
taken over the certified boxes that also give the radius $M$. The input projection, the
convolution and the SiLU bound the change of the scan input $u$. The projections that follow bound
the changes of $B$, of $C$ and of the timescale pre-activation $z$. Lemma~\ref{lem:param}
(Appendix~\ref{app:proofs}) turns the bound on $z$ into $\rho$. A product rule gives the bound
$\eta$ on the change of the equilibrium $c_t = B_t u_t/\lambda$. Theorem~\ref{thm:cell} then
bounds the change of the state in each cell. We carry this bound through the readout
$C_t h_t + D u_t$, the gate and the output projection.

Every stage of the certificate takes the worst case over its box and ignores the dependence
between its inputs. Section~\ref{sec:sens-experiments} bounds from above how much is lost up to
each stage.

\section{Experiments}
\label{sec:experiments}

\subsection{Setup}
\label{sec:exp-setup}

Our data are
the Databento MBP-10 records of SPY on 28~January 2021. After feature construction they give about
$4.55$ million rows and $43$ model inputs. We train a small model on windows of $100$ ticks. From
each window it predicts the log return of the mid-price over the next $100$ ticks.

We divide the records chronologically into a training split of $70$ per cent and validation and
test splits of $15$ per cent each. We purge each boundary by the prediction horizon of $100$
ticks. We winsorise every feature at the $0.001$ and $0.999$ quantiles of its training values, in
training and at inference alike.

The model maps the $43$ inputs to width $d_{\mathrm{model}} = 64$ with a linear layer, a SiLU and a
layer normalisation. Two Mamba blocks follow, each with $d_{\mathrm{inner}} = 128$ channels and
$d_{\mathrm{state}} = 16$ state indices per channel. Each block has a convolution of width $4$ and
a timescale projection of rank $4$. A final layer normalisation and a two-layer head read the
prediction at the last position.

We train two versions of this model, with set-ups that differ only in the timescale map. We call
each of the two trained models an arm. The bounded arm uses the bounded map of
Section~\ref{sec:block} with $\Delta$ in $[10^{-3}, 10^{-1}]$, the range from which Mamba
initialises its timescales \citep{gu2023mamba}. The softplus arm uses Mamba's map
$\Delta = \mathrm{softplus}(z)$. Both arms use the exact zero-order hold~\eqref{eq:zoh}, not the
exponential-Euler input term of the released implementations (Section~\ref{sec:euler}). After
training, the decay rates $\lambda$ of the bounded arm lie between $0.912$ and $17.14$.

We score predictions by the rank information coefficient, the Spearman correlation between
predicted and realised returns. For each arm we keep the checkpoint with the best rank information
coefficient on a fixed slice of $12{,}800$ windows of the validation split. Training ran for
at most $3000$ steps, and the kept checkpoints come from step $750$ on the bounded arm and step
$2000$ on the softplus arm.

Each arm is a single training run, and the two kept checkpoints come from different training
steps, so we draw no conclusion about the timescale maps from differences between the arms. We
keep the softplus arm to show what a very small certified floor does to the additive
baseline.

On the first $25{,}600$ windows of the test split the bounded arm has a rank information
coefficient of $0.13$. Its block
bootstrap $95$ per cent interval is $[0.075, 0.197]$, so this arm has learned some signal. We do
not compare it with other architectures.

We bound how much a perturbation of a block's input changes the block's output before the residual
connection. This output is $F = \mathrm{out\_proj}(y \odot \mathrm{SiLU}(g))$, where $y$ is the
readout and $g$ the gate pre-activation (Appendix~\ref{app:block}). The perturbation moves the
normalised input of the block by at most $\delta$ in sup-norm at every position. We divide each
bound, and each change that an attack finds, by $\delta$ and call the quotient a slope. For a
certificate we take the limit of this quotient as $\delta \to 0$.

When we compare a certificate with an attack, we compute the certificate on boxes widened to cover
every point within $\delta$ of a normalised vector (Appendix~\ref{app:block}). The widening is
needed because an attacked point need not be a normalised vector. At $\delta = 10^{-2}$ it raises
the certificate by less than $5.2$ per cent.

We compute the boxes and radii of the certificate by interval arithmetic in float32, and the rest
of the certificate in round-to-nearest float64. No stage uses outward rounding. The relative
rounding error per operation is of order $10^{-7}$ in float32 and $10^{-16}$ in float64.

Each attack runs sign-gradient ascent in float64, and we re-evaluate the change it finds with a
reference forward pass. Every attack value is therefore a lower bound on the true sensitivity. We
call the largest slope that an attack finds the attack slope. The ratio of a certificate's slope to
the attack slope is an upper bound on the certificate's looseness.

\subsection{One cell}
\label{sec:one-cell}

We test the tightness of Theorem~\ref{thm:cell} on a single cell by searching numerically for the
largest gap between two runs. The cell has $\lambda = 1$ and $M = 1$, and both runs start from
$0$. The gap after $T$ steps is $|h_{T-1} - \tilde h_{T-1}|$. The search runs gradient ascent with
Adam in float64 for $600$ steps from $64$ random initial sequences and from initial sequences
shaped like the construction of Proposition~\ref{prop:lower}. In the shaped sequences, the
perturbed equilibrium $\tilde c_t$ follows a square wave with runs of $1$, $2$, $4$, $8$ or $16$
steps. We re-run
the best sequence found with the plain recursion~\eqref{eq:convex}. Every gap we report is
therefore attained by two admissible runs and is a lower bound on the true worst case.

We first search the hypothesis class of Theorem~\ref{thm:cell}, with the nominal timescale
$\lambda\Delta_t$ free in $[10^{-4}, 10]$. We take $\rho \in \{0.01, 0.03, 0.1, 0.3, 0.6\}$ and
$\eta \in \{0, 0.1\}$ at lengths $T$ from $10$ to $1000$.

No gap found at any searched length exceeds $\eta + 2M\tanh(\rho/2)$, the value that the
construction of Proposition~\ref{prop:lower} approaches. For $T \ge 300$ Adam from the initial sequences shaped
like the construction comes within $0.01$ per cent of this value at every $\rho$ and $\eta$. At the same
lengths, Adam from the $64$ random initial sequences comes within $0.4$ per cent of it. At $T = 10$ the
largest gap found lies less than $1.1$ per cent below the value.

\begin{figure}[htbp]
  \centering
  \includegraphics[width=\textwidth]{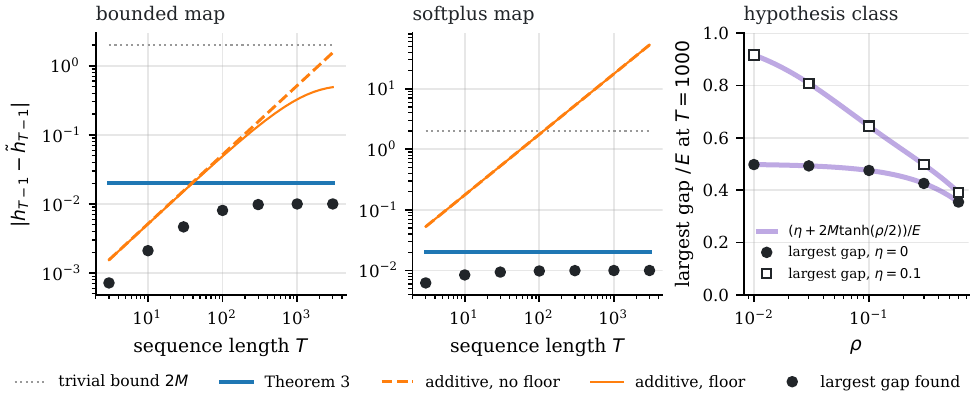}
  \caption{Largest gap found by numerical search in one cell with $\lambda = 1$ and $M = 1$, with
    both runs started from $0$. The left and middle panels show the largest gap found against the
    length $T$ for the bounded and the softplus map, at $\rho = 0.01$ and $\eta = 0$. They also
    show the bound of Theorem~\ref{thm:cell}, the additive baseline~\eqref{eq:additive} with and
    without the floor $\Delta(\inf Z)$, and the trivial bound $2M + \eta = 2$. The gap levels off at
    about half of the theorem's bound. On the softplus map the two forms of the baseline coincide
    at every length shown. The right panel shows the largest gap found at $T = 1000$, divided by the
    theorem's bound $E$, against $\rho$ for $\eta = 0$ and $\eta = 0.1$. The curve for each $\eta$ is the value
    $\eta + 2M\tanh(\rho/2)$ that the construction of Proposition~\ref{prop:lower} approaches,
    divided by $E$. Every point lies less than $0.01$ per cent below the curve, and none lies
    above it.}
  \label{fig:one-cell}
\end{figure}

For $\eta = 0$ and $T \ge 300$, the theorem's bound $E = \eta + 2M(e^{\rho} - 1)$ equals $e^{\rho} + 1$ times the largest gap found, to four significant figures. This factor
is about $2.0$ at $\rho = 0.01$ and about $2.8$ at $\rho = 0.6$ (Figure~\ref{fig:one-cell},
right).

We then set $\eta = 0$ and search the perturbation model of our certificates. In this model each
timescale is the image of a pre-activation under the timescale map. The nominal pre-activation
$z_t$ lies in a box $Z$, and the perturbed one lies in $Z$ within $\zeta$ of $z_t$. For the
bounded map we take $\Delta \in [10^{-3}, 10^{-1}]$ and $Z = [-8, 8]$. For the softplus map we
take $Z = [-48, 2]$. On both maps we choose $\zeta$ so that the bound on the change of $\log\Delta$
is $\rho = \zeta\sup_Z|(\log\Delta)'| = 0.01$. The additive baseline~\eqref{eq:additive} with
$L = T$ uses the bound $\eta_\Delta = \zeta\sup_Z|\Delta'|$ on the change of $\Delta$, and its
floored form uses the floor $\Delta_{\mathrm{lo}} = \Delta(\inf Z)$. We search at lengths $T$ from
$3$ to $3000$.

On both maps the largest gap found grows with $T$ and levels off at about half of the theorem's
bound (Figure~\ref{fig:one-cell}, left and middle). This is the value that the search over the
hypothesis class gives for $\rho = 0.01$ and $\eta = 0$.

Since $M = 1$ and $\eta = 0$ here, every gap between two admissible runs is at most the
trivial bound $2M + \eta = 2$ of Section~\ref{sec:sensitivity}. This bound does not tend to zero
with the perturbation, so it gives no finite slope and plays no part in the block certificates of
Section~\ref{sec:floor-exp}.

For the bounded map the additive baseline is above the theorem's bound from $T = 39$ without the
floor and from $T = 40$ with it. Both lengths are about inversely proportional to $\lambda$. With
the same box and $\zeta$, at $\lambda = 17.14$, the largest decay rate of the trained bounded arm,
the baseline would be above the theorem's bound from $T = 3$. Without the floor the baseline keeps
growing linearly. With the floor it
tends to about $25.6$ times the theorem's bound as $T \to \infty$, below the value $25.8$ of
Remark~\ref{rem:cap}. Both forms of the baseline stay below the trivial bound at every searched
length.

For the softplus map the floor $\Delta(\inf Z)$ is about $1.4 \times 10^{-21}$. At every searched
length the baseline with this floor coincides with the one without it. The baseline is above the
theorem's bound from $T = 2$ and above the trivial bound from $T = 114$. It grows linearly to more
than $2600$ times the theorem's bound at $T = 3000$.

\subsection{Comparison with the additive baseline}
\label{sec:floor-exp}

\begin{figure}[htbp]
  \centering
  \includegraphics[width=\textwidth]{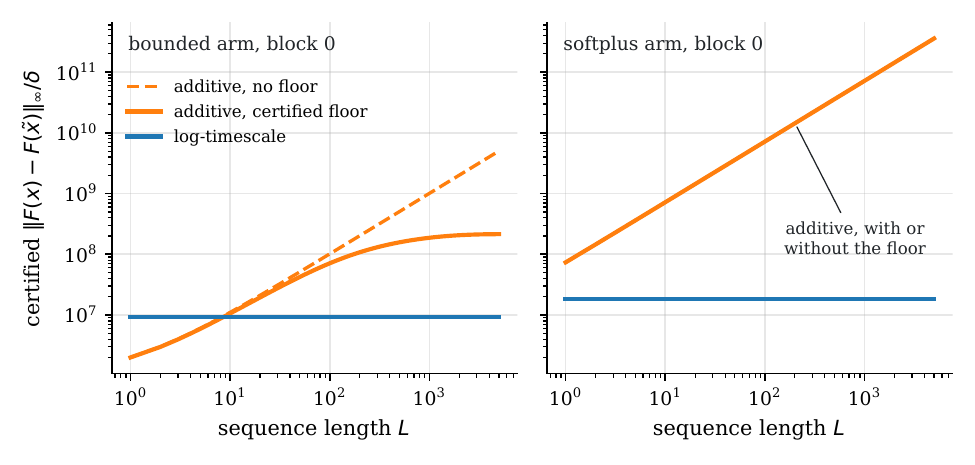}
  \caption{Data-free output slopes of block~0 of each arm against the sequence length $L$. The
    flat line, labelled log-timescale, is our certificate, which uses the constant $E$ of
    Theorem~\ref{thm:cell} in every cell. Without a floor the additive
    baseline~\eqref{eq:additive} grows linearly, and with the certified floor it levels off as
    $L \to \infty$. On the bounded arm both forms of the baseline pass the flat line near
    $L = 8$. On the softplus arm the smallest
    certified floor is about $10^{-21}$, and the two forms of the baseline overlap at every
    length shown.}
  \label{fig:vs-length}
\end{figure}

Table~\ref{tab:incremental} and Figure~\ref{fig:vs-length} compare our data-free certificate with
one that uses the additive baseline~\eqref{eq:additive} in place of Theorem~\ref{thm:cell} in each
cell. Our certificate does not depend on the length $L$, while the baseline without a floor grows
linearly with $L$. On the bounded arm the baseline passes our certificate after about $8$ steps,
the crossing $L$ in Table~\ref{tab:incremental}. It grows by about $0.11$ times our certificate
per step. At $L = 5000$ it is about $540$ times ours on both blocks. On the softplus
arm the baseline is looser at every length. It grows by about $3.9$ times our certificate per
step, and at $L = 5000$ it is about $2 \times 10^{4}$ times ours. These ratios at $L = 5000$ depend
on the chosen length, whereas the crossing $L$ and the growth per step are properties of the
trained model.

Our certificate, the flat line in
Figure~\ref{fig:vs-length}, uses the constant $E$ of the theorem in every cell. The theorem also
gives the finite-horizon bound $(1 - P_t)E$, which is smaller than $E$, so the baseline passes
this bound no later than it passes $E$. We also compare the two time-warp terms cell by cell, with
our finite-horizon term at its worst over admissible timescales. Over the cells of the bounded
arm, the baseline's term reaches ours after a median of $5$ steps ($n^{\ast}$ in
Appendix~\ref{app:cao})\@.

With the certified floor the baseline levels off as $L \to \infty$. On the bounded arm its limit
is $23$ times our certificate on block~0 and $22$ times on block~1. This ratio is set mainly by
our choice of $\Delta_{\max}/\Delta_{\min} = 100$ for the bounded map. On the bounded arm every
certified floor is below $1.1 \times 10^{-3}$ and every certified ceiling is above $0.095$. Since
the bounded map is increasing and gives $\Delta = 10^{-2}$ at $z = 0$, every box $Z$ contains $0$.
Remark~\ref{rem:cap} then bounds the ratio of the two time-warp terms in each cell by $25.8$, a
cap that depends only on $\Delta_{\max}/\Delta_{\min}$. Over the
cells of the trained model the ratio lies between $24.1$ and $25.8$.

The ratio for the block is slightly below the ratio for a single cell. The two certificates
differ only in the time-warp term, and the terms they share make up $0.11$ to $0.13$ of ours at
its largest output coordinate.

With a floor taken from an attack on the block's input, the baseline on the bounded arm is still
$8.5$ to $12$ times our certificate (Appendix~\ref{app:cao})\@. This floor, which we call the
empirical floor, is the smallest timescale that the attack reaches in a channel. Any valid floor
of a channel is at most its empirical floor, and a larger floor lowers the baseline. The empirical
floor is therefore at least as favourable to the baseline as any valid floor.

On the softplus arm the certified floors are too small to matter at the lengths we test. The
smallest are $1.3 \times 10^{-21}$ on block~0 and $1.9 \times 10^{-17}$ on block~1. Up to
$L = 5000$ the baseline with these floors stays within $1$ per cent of the baseline without one.
Its limit as $L \to \infty$ is $2.1 \times 10^{19}$ times our certificate on block~0 and
$2.8 \times 10^{15}$ times on block~1.

\begin{table}[htbp]
\centering
\small
\setlength{\tabcolsep}{4pt}
\begin{tabular}{@{}llrrrrrr@{}}
\toprule
Arm & Block & Theorem~\ref{thm:cell} & \shortstack{Additive, no floor,\\ $L = 5000$} & Ratio & \shortstack{Additive, floor,\\ $L \to \infty$} & Ratio & \shortstack{Crossing\\ $L$} \\
\midrule
bounded  & 0 & $9.34 \times 10^{6}$ & $5.04 \times 10^{9}$  & $539$                & $2.16 \times 10^{8}$  & $23.1$               & $8.3$ \\
bounded  & 1 & $6.08 \times 10^{6}$ & $3.27 \times 10^{9}$  & $538$                & $1.37 \times 10^{8}$  & $22.4$               & $8.1$ \\
softplus & 0 & $1.85 \times 10^{7}$ & $3.59 \times 10^{11}$ & $1.95 \times 10^{4}$ & $3.93 \times 10^{26}$ & $2.1 \times 10^{19}$ & none \\
softplus & 1 & $1.46 \times 10^{7}$ & $2.87 \times 10^{11}$ & $1.97 \times 10^{4}$ & $4.11 \times 10^{22}$ & $2.8 \times 10^{15}$ & none \\
\bottomrule
\end{tabular}
\caption{Data-free output slopes of each block of the two arms, maximised over output coordinates
  and rounded up. The additive columns use the baseline~\eqref{eq:additive} without a floor at
  $L = 5000$ and with the certified floor as $L \to \infty$. Each ratio divides the unrounded
  additive slope to its left by the unrounded slope in the Theorem~\ref{thm:cell} column. The
  crossing $L$ is the length at which the affine baseline without a floor reaches the flat bound
  of Theorem~\ref{thm:cell}. A crossing of none means that the baseline is the larger bound at every length.}
\label{tab:incremental}
\end{table}

\subsection{Comparison with gradient attacks}
\label{sec:sens-experiments}

We attack each block on four held-out windows at each of the lengths $L = 10$, $100$, $1000$ and
$5000$. The windows of all four lengths end at the same rows, and the two longest are $10$ and
$50$ times as long as the training windows. From $L = 10$ to
$L = 5000$ at $\delta = 10^{-6}$, the attack slope grows by a factor of $1.4$ to $7.3$, while the
baseline without a floor grows by a factor of at least $445$. The largest change found lies at the
last position in only $1$ of the $16$ combinations of arm, block and length. At the last position
alone, the attack slope grows by a factor of $1.06$ to $2.3$. The growth
therefore comes partly from searching more positions and partly from the longer history before a
single position.

No attack value exceeds a certificate, and our data-free certificate lies far above what the
attacks find. At $\delta = 10^{-6}$ it is $3.06 \times 10^{5}$ to $3.23 \times 10^{6}$ times the
attack slope. We also compare the certified bound after each stage of the block with the largest
change that random perturbations of the block input produce there. Random perturbations do not
search for the worst case, so each such ratio is an upper bound on the looseness after its stage.
On block~0 of the bounded arm the ratio is $8.2$ after the scan input $u$, $134$ after $B$, $537$
after the equilibrium $c$, $1.54 \times 10^{4}$ after the state $h$, $3.24 \times 10^{5}$ after the
readout $y$ and $4.65 \times 10^{7}$ at the output.
At the output this ratio exceeds the ratio to the attack slope.

The factor of a stage is the ratio after it divided by the ratio before it. In every block the
gate and the output projection together have the largest factor, $98$ to $147$. The scan follows
with $18$ to $39$, and then the readout with $14$ to $29$. The step that forms the equilibrium
from $B_t$ and $u_t$ has a factor of only $4$ to $7$. Each factor mixes the slack of its stage
with how far the random samples fall short of that stage's worst case.

Most of this slack comes from taking the worst case over the certified boxes. Around the realised
input of each window, we also propagate the perturbation through every stage of the block. In the
scan we apply the triangle inequality to the error identity~\eqref{eq:error} with the two runs
exchanged. The realised distance $|h_{t-1} - c_t|$ then takes the place of the bound $2M$ that the
proof uses, and we take the result at the worst transition that the perturbed run can have. That
transition lies in the intersection of the intervals given by the log-timescale bound and by the
additive bound on $|\Delta_t - \tilde\Delta_t|$. For $\delta \le 10^{-4}$ this local bound is at
most $12.7$ times the attack slope, and dropping the log-timescale interval from the intersection
changes it by at most $0.034$ per cent. The local bound and its code are in our repository.

To first order in the perturbation, the choice between the log and the additive metric matters
only for the data-free certificate. A bound on the change of the timescale pre-activation $z_t$
bounds the change of $\Delta_t$ through the derivative $\Delta'$ and that of $\log\Delta_t$ through
$(\log\Delta)'$. Since $\Delta' = \Delta\,(\log\Delta)'$, the first bound is $\Delta_t$ times the
second to first order. Along a given input the realised timescale $\Delta_t$ is known at every
step, so either bound gives the other and the two agree to first order. The data-free certificate
has no realised timescale to use, and there the two metrics give the different certificates of
Section~\ref{sec:floor-exp}.

\section{Related work}
\label{sec:related}

\paragraph{Bounds on selective scans.}
Stability analyses of selective scans differ in how they treat the input-dependent timescale.
Halloran, Gulati and Roysdon \citep{halloran2025lyapunov} show that the maximal Lyapunov exponent
of the Mamba recurrence is nonpositive when $\Delta_t$, $B_t$ and $C_t$ are held at their nominal
values. Zubi\'{c} and Scaramuzza \citep{zubic2026selective} work in continuous time under a common
quadratic Lyapunov function. Cao et al.~\citep{cao2026control} keep the timescale input-dependent
and treat one layer without normalisation, for inputs in a known bounded set.
Section~\ref{sec:compare} compares their bound with ours.

Analyses of training and generalisation for selective state space models also rely on a floor.
Piaseczny et al.~\citep{piaseczny2026distributed} bound parameter gradients uniformly in the length
under a floor on the softplus timescale. The robustness bound of Tang et
al.~\citep[Theorem~1]{tang2026selective} takes its decay factor from a floor on the timescale.

\paragraph{Incremental stability of gated recurrences.}
A factor that plays the role of a floor also appears in incremental stability results for gated
recurrent networks. The incremental stability conditions of Bonassi et al.~\citep{bonassi2021gru}
for networks of gated recurrent units contain the inverse of one minus an upper bound on the
retention gate. The incremental gains of De Carli et al.~\citep{decarli2026dgn,decarli2026mgu}
contain the inverse of one minus a contraction bound, which includes an upper bound on the forget
gate.

Orvieto et al.~\citep[Section~4]{orvieto2023lru} observe that the factor $\Delta$ in the input gain
cancels the growth of order $1/\Delta$ that the transition alone would produce. Zucchet and Orvieto
\citep{zucchet2024recurrent} use logarithmic timescale coordinates to control the parameter
sensitivity of diagonal linear recurrences. Their coordinates describe the fixed eigenvalues of the
recurrence, whereas the timescale of our cell changes at every step with the input.

The nearest argument we know is that of Matsubara
et al.~\citep[Lemma~A.5]{matsubara2026exponential}. They bound the error between two exponential
smoothers through a recursion like~\eqref{eq:error}, with a constant proportional to the relative
difference of their smoothing parameters. Each cell is such a smoother on the irregular grid of
accumulated times \citep[Section~4.1]{eckner2019algorithms}, but its input weight varies from step
to step. Lemma~\ref{lem:ratio} bounds the difference of the
two transitions at each step by a multiple of that step's input weight, which keeps these
differences from accumulating.

\paragraph{Verification of recurrent models.}
Verifiers for recurrent networks either propagate an input set through a fixed number of steps or
search for an invariant. Propagation certifies one length at a time, whether it unrolls the network
into a feed-forward one or proceeds step by step with star sets \citep{tran2023star}. A search for
an inductive invariant may iterate the transition up to a depth limit \citep{hosseini2025ltl} or
check candidates with Marabou \citep{jacoby2020rnn,katz2019marabou}. It succeeds only if it finds
an invariant. For state space models, Thapliyal \citep{thapliyal2026contraction} propagates
intervals exactly through a linear time-invariant classifier and observes that the analysis does
not carry over to Mamba.

Brucker and Stell \citep{brucker2023isabelle} give Isabelle/HOL proofs for feed-forward
classifiers, and George et al.~\citep{george2026torchlean} give Lean proofs for unrolled recurrent
computation. Our Isabelle theory is narrower than both.

\section{Limitations}
\label{sec:limitations}

Theorem~\ref{thm:cell} concerns one scalar cell, so it requires a diagonal $A$. It also needs a
real transition $\Abar_t$ in $(0, 1)$, which makes each step a convex combination. The theorem
therefore does not cover the complex-valued state update of Mamba-3
\citep[Section~3.2]{lahoti2026mamba3}, whose transition is not a real number in $(0, 1)$.

We do not prove that the largest gap from a common start equals $\eta + 2M\tanh(\rho/2)$. The
search of Section~\ref{sec:one-cell} gives only lower bounds on this gap. For $\eta > 0$, the
construction of Proposition~\ref{prop:lower} puts the nominal equilibrium outside $[-M, M]$, at
$c_t = M + \eta$ on half of the steps. When the equilibria of both runs are restricted to
$[-M, M]$, so are both states, and the gap is at most $2M$. Under this restriction no gap can
approach $\eta + 2M\tanh(\rho/2)$ once $\eta > 4M/(e^{\rho} + 1)$, because this value then exceeds
$2M$. With $\tilde c_t$ alternating between $M - \eta$ and $-M$ instead, the construction satisfies
the restriction and gives gaps that approach $\eta + (2M - \eta)\tanh(\rho/2)$. We do not know the
largest gap under this restriction.

Our certificates bound one block at a time, with respect to its normalised input. We do not compose
them through the residual stream and the head. A certificate with respect to the raw input of a
block would also have to pass through its layer normalisation (Lemma~\ref{lem:ln}). This
normalisation is globally Lipschitz because its variance offset $\varepsilon$ is positive. In the
sup-norm of our certificates, its Jacobian on the blocks of the bounded arm has norm at most $3.9$
at realised inputs but about $653$ at a constant input. Because a certificate holds for every
input, one with respect to the raw input would carry the global Lipschitz constant, which is at
least the norm of about $653$ at a constant input. We do not compute one.

Theorem~\ref{thm:cell} and its corollaries hold in real arithmetic, and we evaluate our
certificates without outward rounding (Section~\ref{sec:exp-setup}). A scan computed in float32
adds a rounding error at every step. In a cell this error decays at the rate $\Abar$, so it
accumulates to at most about $1/(1 - \sup\Abar)$ times the error added per step. On the bounded
arm this factor is about $1100$. One run there accumulates an error of at most about
$2 \times 10^{-4} M$, where $M$ is the invariant radius of the cell. At $\delta = 10^{-6}$, the
errors of the two runs together exceed the certified change of the state in $49$ and $61$ per cent
of the cells of the two blocks. A float32 scan is therefore
covered only once a rounding term is added to the bound.

A transition can also round to exactly $1$. In bfloat16 this happens to the slowest transitions of
the bounded arm. In float32 it happens to the slowest transitions of the softplus arm, where
$\lambda\Delta$ reaches $10^{-21}$ on the certified boxes. Where this happens, the computed update
is not a convex combination, and the theorem does not apply to the scan as computed.

Our order book data come from one instrument on one trading day, and our attacks are heuristic
searches on four windows. The two arms come from one training set-up with one learning rate. They
use the exact zero-order hold rather than the exponential-Euler step of the released
implementations. We do not certify a released language model here.

\section{Conclusion}
\label{sec:conclusion}

We have shown that measuring timescale perturbations in $\log\Delta$ rather than in $\Delta$ gives
a sensitivity bound for a selective-scan cell that holds at every sequence length and needs no
floor on the timescale (Theorem~\ref{thm:cell}). From a common start, its constant is within a
factor $e^{\rho} + 1$ of the best possible (Proposition~\ref{prop:lower}).

How much this helps in practice depends on the timescale map. With the bounded map, the additive
baseline is tighter than our certificate only over the first $8$ steps or fewer, and once it uses
its certified floor, the chosen timescale range caps its ratio to ours in each cell
(Remark~\ref{rem:cap}). With Mamba's softplus map, the certified floor is the only thing that
limits the baseline, and on our model it is too small to matter at any length we test. Along a
single input, the log and additive ways of measuring the change agree to first order, so the gain
shows up only in bounds that must hold for every input of a block.

This paper does not settle whether $\eta + 2M\tanh(\rho/2)$ is the largest possible gap from a
common start, although our search finds none larger, nor what the largest gap is when the
equilibria of both runs lie in $[-M, M]$ (Section~\ref{sec:limitations}). Natural next steps are
to extend the theorem to the complex-valued transitions of Mamba-3 and to tighten the block
certificate, which remains far above what attacks find.

\section*{Use of generative AI}

Claude (Anthropic), a large language model, assisted the authors throughout this project. It
assisted with searching and summarising the related literature, drafting and checking proofs,
writing and debugging code, running experiments, and drafting and revising the text. It was
also used to review the paper, including tracing every reported number to the code and the
recorded runs in our repository. The authors reviewed all AI-assisted material and take full
responsibility for the content of this paper.

\bibliographystyle{plainnat}
\bibliography{references}

@inproceedings{gu2023mamba,
  author    = {Gu, Albert and Dao, Tri},
  title     = {Mamba: Linear-Time Sequence Modeling with Selective State Spaces},
  booktitle = {First Conference on Language Modeling (COLM)},
  year      = {2024},
  note      = {arXiv:2312.00752},
}

@inproceedings{jacoby2020rnn,
  author    = {Jacoby, Yuval and Barrett, Clark and Katz, Guy},
  title     = {Verifying Recurrent Neural Networks Using Invariant Inference},
  booktitle = {Automated Technology for Verification and Analysis (ATVA)},
  series    = {Lecture Notes in Computer Science},
  volume    = {12302},
  pages     = {57--74},
  year      = {2020},
  publisher = {Springer},
  note      = {Full version: arXiv:2004.02462},
}

@inproceedings{katz2019marabou,
  author    = {Katz, Guy and Huang, Derek A. and Ibeling, Duligur and Julian, Kyle
               and Lazarus, Christopher and Lim, Rachel and Shah, Parth and
               Thakoor, Shantanu and Wu, Haoze and Zelji{\'c}, Aleksandar and
               Dill, David L. and Kochenderfer, Mykel J. and Barrett, Clark},
  title     = {The {Marabou} Framework for Verification and Analysis of Deep Neural Networks},
  booktitle = {Computer Aided Verification (CAV)},
  series    = {Lecture Notes in Computer Science},
  volume    = {11561},
  pages     = {443--452},
  year      = {2019},
  publisher = {Springer},
}

@inproceedings{tran2023star,
  author    = {Tran, Hoang-Dung and Choi, Sung Woo and Yang, Xiaodong and
               Yamaguchi, Tomoya and Hoxha, Bardh and Prokhorov, Danil},
  title     = {Verification of Recurrent Neural Networks with Star Reachability},
  booktitle = {Hybrid Systems: Computation and Control (HSCC)},
  year      = {2023},
  publisher = {ACM},
  doi       = {10.1145/3575870.3587128},
}

@article{bonassi2021gru,
  author  = {Bonassi, Fabio and Farina, Marcello and Scattolini, Riccardo},
  title   = {On the Stability Properties of Gated Recurrent Units Neural Networks},
  journal = {Systems \& Control Letters},
  volume  = {157},
  pages   = {105049},
  year    = {2021},
  doi     = {10.1016/j.sysconle.2021.105049},
}

@book{nipkow2002isabelle,
  author    = {Nipkow, Tobias and Paulson, Lawrence C. and Wenzel, Markus},
  title     = {Isabelle/HOL: A Proof Assistant for Higher-Order Logic},
  series    = {Lecture Notes in Computer Science},
  volume    = {2283},
  publisher = {Springer},
  year      = {2002},
}

@inproceedings{cao2026control,
  author    = {Cao, Liang and Liu, Weide and Chen, Zhuo and Qin, Yan},
  title     = {A Control-Theoretic View of {Mamba} on Stability and Robustness},
  booktitle = {Proceedings of the 43rd International Conference on Machine Learning},
  series    = {Proceedings of Machine Learning Research},
  volume    = {306},
  pages     = {11350--11364},
  publisher = {PMLR},
  year      = {2026},
}

@article{horbatko2026sessa,
  author  = {Horbatko, Liubomyr},
  title   = {{Sessa}: Selective State Space Attention},
  journal = {arXiv preprint arXiv:2604.18580},
  year    = {2026},
}

@article{zubic2026selective,
  author  = {Zubi{\'c}, Nikola and Scaramuzza, Davide},
  title   = {Regularity and Stability Properties of Selective {SSMs} with Discontinuous Gating},
  journal = {Transactions on Machine Learning Research},
  issn    = {2835-8856},
  year    = {2026},
  url     = {https://openreview.net/forum?id=7Vav53cDeN},
  note    = {arXiv:2505.11602},
}

@article{halloran2025lyapunov,
  author  = {Halloran, John T. and Gulati, Manbir and Roysdon, Paul F.},
  title   = {{Mamba} State-Space Models Are {Lyapunov}-Stable Learners},
  journal = {Transactions on Machine Learning Research},
  issn    = {2835-8856},
  year    = {2025},
  url     = {https://openreview.net/forum?id=wzsYQYs3dO},
  note    = {arXiv:2406.00209},
}

@inproceedings{hosseini2025ltl,
  author    = {Hosseini, Mehran and Lomuscio, Alessio and Paoletti, Nicola},
  title     = {{LTL} Verification of Memoryful Neural Agents},
  booktitle = {Proceedings of the 24th International Conference on Autonomous Agents
               and Multiagent Systems (AAMAS 2025)},
  pages     = {978--987},
  publisher = {International Foundation for Autonomous Agents and Multiagent Systems},
  address   = {Detroit, Michigan, USA},
  year      = {2025},
  note      = {arXiv:2503.02512},
}

@inproceedings{lahoti2026mamba3,
  author    = {Lahoti, Aakash and Li, Kevin Y. and Chen, Berlin and Wang, Caitlin and
               Bick, Aviv and Kolter, J. Zico and Dao, Tri and Gu, Albert},
  title     = {{Mamba-3}: Improved Sequence Modeling using State Space Principles},
  booktitle = {International Conference on Learning Representations (ICLR)},
  year      = {2026},
  url       = {https://openreview.net/forum?id=HwCvaJOiCj},
  note      = {arXiv:2603.15569},
}

@inproceedings{gu2021lssl,
  author    = {Gu, Albert and Johnson, Isys and Goel, Karan and Saab, Khaled and
               Dao, Tri and Rudra, Atri and R{\'e}, Christopher},
  title     = {Combining Recurrent, Convolutional, and Continuous-time Models with
               Linear State Space Layers},
  booktitle = {Advances in Neural Information Processing Systems},
  volume    = {34},
  pages     = {572--585},
  publisher = {Curran Associates, Inc.},
  year      = {2021},
  note      = {arXiv:2110.13985},
}

@inproceedings{tallec2018warp,
  author    = {Tallec, Corentin and Ollivier, Yann},
  title     = {Can recurrent neural networks warp time?},
  booktitle = {International Conference on Learning Representations (ICLR)},
  year      = {2018},
  url       = {https://openreview.net/forum?id=SJcKhk-Ab},
  note      = {arXiv:1804.11188},
}

@inproceedings{orvieto2023lru,
  author    = {Orvieto, Antonio and Smith, Samuel L. and Gu, Albert and Fernando, Anushan and
               Gulcehre, Caglar and Pascanu, Razvan and De, Soham},
  title     = {Resurrecting Recurrent Neural Networks for Long Sequences},
  booktitle = {Proceedings of the 40th International Conference on Machine Learning},
  series    = {Proceedings of Machine Learning Research},
  volume    = {202},
  pages     = {26670--26698},
  publisher = {PMLR},
  year      = {2023},
  note      = {arXiv:2303.06349},
}

@article{thapliyal2026contraction,
  author  = {Thapliyal, Omanshu},
  title   = {Bounded Reachability \& Jailbreak Detection via Contraction-Constrained
             State Space Models},
  journal = {arXiv preprint arXiv:2610.02853},
  year    = {2026},
  note    = {Accepted at the First Workshop in AI Measurement Science (AIMS), COLM 2026},
}

@article{decarli2026dgn,
  author  = {De Carli, Stefano and Previtali, Davide and Mazzoleni, Mirko and Previdi, Fabio},
  title   = {Chaos-Free Networks Are Stable Recurrent Neural Networks},
  journal = {IEEE Control Systems Letters},
  volume  = {10},
  pages   = {751--756},
  year    = {2026},
  doi     = {10.1109/LCSYS.2026.3702800},
  note    = {arXiv:2603.14106},
}

@article{decarli2026mgu,
  author  = {De Carli, Stefano and Previtali, Davide and Mazzoleni, Mirko and Previdi, Fabio},
  title   = {Stability properties of {Minimal Gated Unit} neural networks},
  journal = {arXiv preprint arXiv:2603.03017},
  year    = {2026},
}

@article{tang2026selective,
  author  = {Tang, Alex and Ildiz, M. Emrullah and Kurt, Batin and Oymak, Samet and Ozay, Necmiye},
  title   = {On the Generalization Properties of Selective State-Space Models for
             Filtering Tasks for Unknown Systems},
  journal = {arXiv preprint arXiv:2604.23818},
  year    = {2026},
  note    = {Accepted to the 65th IEEE Conference on Decision and Control (CDC 2026)},
}

@inproceedings{zucchet2024recurrent,
  author    = {Zucchet, Nicolas and Orvieto, Antonio},
  title     = {Recurrent neural networks: vanishing and exploding gradients are not
               the end of the story},
  booktitle = {Advances in Neural Information Processing Systems},
  volume    = {37},
  pages     = {139402--139443},
  publisher = {Curran Associates, Inc.},
  year      = {2024},
  doi       = {10.52202/079017-4425},
  note      = {arXiv:2405.21064},
}

@article{samuelson1968deviant,
  author  = {Samuelson, Paul A.},
  title   = {How Deviant Can You Be?},
  journal = {Journal of the American Statistical Association},
  volume  = {63},
  number  = {324},
  pages   = {1522--1525},
  year    = {1968},
  doi     = {10.1080/01621459.1968.10480944},
}

@article{laguerre1880methode,
  author  = {Laguerre, Edmond},
  title   = {Sur une m{\'e}thode pour obtenir par approximation les racines d'une
             {\'e}quation alg{\'e}brique qui a toutes ses racines r{\'e}elles},
  journal = {Nouvelles Annales de Math{\'e}matiques, 2e s{\'e}rie},
  volume  = {19},
  pages   = {161--171},
  year    = {1880},
  url     = {https://www.numdam.org/item/NAM_1880_2_19__161_1/},
}

@article{riechers2024layernorm,
  author  = {Riechers, Paul M.},
  title   = {Geometry and Dynamics of {LayerNorm}},
  journal = {arXiv preprint arXiv:2405.04134},
  year    = {2024},
}

@inproceedings{brody2023layernorm,
  author    = {Brody, Shaked and Alon, Uri and Yahav, Eran},
  title     = {On the Expressivity Role of {LayerNorm} in Transformers' Attention},
  booktitle = {Findings of the Association for Computational Linguistics: ACL 2023},
  pages     = {14211--14221},
  address   = {Toronto, Canada},
  publisher = {Association for Computational Linguistics},
  year      = {2023},
  doi       = {10.18653/v1/2023.findings-acl.895},
  note      = {arXiv:2305.02582},
}

@inproceedings{brucker2023isabelle,
  author    = {Brucker, Achim D. and Stell, Amy},
  title     = {Verifying Feedforward Neural Networks for Classification in {Isabelle/HOL}},
  booktitle = {Formal Methods: 25th International Symposium, FM 2023},
  series    = {Lecture Notes in Computer Science},
  volume    = {14000},
  pages     = {427--444},
  publisher = {Springer},
  year      = {2023},
  doi       = {10.1007/978-3-031-27481-7_24},
}

@article{george2026torchlean,
  author  = {George, Robert Joseph and Cruden, Jennifer and Adkisson, Will and
             Zhong, Xiangru and Zhang, Huan and Anandkumar, Anima},
  title   = {{TorchLean}: Formalizing Neural Networks in {Lean}},
  journal = {arXiv preprint arXiv:2602.22631},
  year    = {2026},
}

@book{billingsley1999convergence,
  author    = {Billingsley, Patrick},
  title     = {Convergence of Probability Measures},
  edition   = {Second},
  series    = {Wiley Series in Probability and Statistics},
  publisher = {Wiley},
  address   = {New York},
  year      = {1999},
  isbn      = {978-0-471-19745-4},
  doi       = {10.1002/9780470316962},
}

@article{lohmiller1998contraction,
  author  = {Lohmiller, Winfried and Slotine, Jean-Jacques E.},
  title   = {On Contraction Analysis for Non-linear Systems},
  journal = {Automatica},
  volume  = {34},
  number  = {6},
  pages   = {683--696},
  year    = {1998},
  doi     = {10.1016/S0005-1098(98)00019-3},
}

@book{bullo2026contraction,
  author    = {Bullo, Francesco},
  title     = {Contraction Theory for Dynamical Systems},
  note      = {Edition 1.3},
  publisher = {Kindle Direct Publishing},
  year      = {2026},
  isbn      = {979-8836646806},
  url       = {https://fbullo.github.io/ctds},
}

@article{piaseczny2026distributed,
  author        = {Piaseczny, Adam and Shisher, Md Kamran Chowdhury and Wang, Shiqiang and Brinton, Christopher G.},
  title         = {Distributed Learning with Selective State Space Models: Architecture-Aware Convergence Analysis},
  journal       = {arXiv preprint arXiv:2610.02659},
  year          = {2026},
  eprint        = {2610.02659},
  archivePrefix = {arXiv}
}

@misc{eckner2019algorithms,
  author      = {Eckner, Andreas},
  title       = {Algorithms for Unevenly Spaced Time Series: Moving Averages and Other Rolling Operators},
  howpublished = {Working paper},
  year        = {2019},
  month       = apr,
  url         = {http://www.eckner.com/papers/Algorithms%20for%20Unevenly%20Spaced%20Time%20Series.pdf},
  note        = {First version January 2010; current version 10 April 2019}
}

@article{matsubara2026exponential,
  author        = {Matsubara, Takuo and Jiang, Peiwen and Chen, Wilson Ye and Tran, Minh-Ngoc},
  title         = {Exponential Smoothing for Time Series of Random Objects},
  journal       = {arXiv preprint arXiv:2609.20274},
  year          = {2026},
  eprint        = {2609.20274},
  archivePrefix = {arXiv}
}

\appendix

\section{The lower bound, the corollaries and their proofs}
\label{app:proofs}

\subsection{Proof of the lower bound}
\label{app:lower}

\begin{proof}[Proof of Proposition~\ref{prop:lower}]
Put $Q = e^{\rho} > 1$, fix $\tau > 0$ and put $a = e^{-Q\tau}$ and $b = e^{-\tau}$, so that
$0 < a < b < 1$. On the even steps we take $\tilde c_t = M$, $\lambda\Delta_t = Q\tau$ and
$\tilde\Delta_t = \Delta_t/Q$, and on the odd steps $\tilde c_t = -M$, $\lambda\Delta_t = \tau$
and $\tilde\Delta_t = Q\Delta_t$. On every step $c_t = \tilde c_t + \eta$. Then
$|\log\Delta_t - \log\tilde\Delta_t| = \rho$, $|\tilde c_t| = M$ and $c_t - \tilde c_t = \eta$
for every $t$. The nominal transitions are $a, b, a, b, \dots$ and the perturbed ones
$b, a, b, a, \dots$, starting at $t = 0$.

Let $\hat h$ be the run on the nominal timescales with the equilibria of the perturbed run,
$\hat h_t = \Abar_t\hat h_{t-1} + (1 - \Abar_t)\tilde c_t$ with $\hat h_{-1} = 0$, and split
$h_t - \tilde h_t = (h_t - \hat h_t) + (\hat h_t - \tilde h_t)$. The first part
$d_t = h_t - \hat h_t$ satisfies $d_t = \Abar_t d_{t-1} + (1 - \Abar_t)\eta$ with $d_{-1} = 0$,
so $\eta - d_t = \Abar_t(\eta - d_{t-1})$ and $d_t = \eta(1 - P_t)$, with $P_t$ as in
Theorem~\ref{thm:cell}. Since $P_t \le b^{t+1}$, $d_t$ tends to $\eta$.

For the second part we put $X_k = \hat h_{2k}$ and $\tilde X_k = \tilde h_{2k}$ for $k \ge 0$, so
that $X_0 = (1 - a)M$ and $\tilde X_0 = (1 - b)M$. Step $2k - 1$ has $\tilde c = -M$ and step $2k$
has $\tilde c = M$, so for $k \ge 1$
\[
  X_k = a\bigl(b\, X_{k-1} - (1 - b)M\bigr) + (1 - a)M = ab\, X_{k-1} + M(1 - 2a + ab),
\]
and in the same way, with the two transitions exchanged,
$\tilde X_k = b\bigl(a\, \tilde X_{k-1} - (1 - a)M\bigr) + (1 - b)M = ab\, \tilde X_{k-1} + M(1 - 2b + ab)$. Since
$0 < ab < 1$, both sequences converge to the fixed points of these affine maps, and
\[
  \lim_{k \to \infty} \bigl(\hat h_{2k} - \tilde h_{2k}\bigr)
    = \frac{M(1 - 2a + ab) - M(1 - 2b + ab)}{1 - ab}
    = 2M\,\frac{e^{-\tau} - e^{-Q\tau}}{1 - e^{-(Q + 1)\tau}} .
\]
As $\tau \to 0$ the numerator of the last fraction is $(Q - 1)\tau + O(\tau^2)$ and its
denominator $(Q + 1)\tau + O(\tau^2)$, so the limit tends to
$2M(Q - 1)/(Q + 1) = 2M\tanh(\rho/2)$.

Given $\varepsilon > 0$, we choose $\tau$ such that $2M(b - a)/(1 - ab) > 2M\tanh(\rho/2) -
\varepsilon/2$. Since $h_{2k} - \tilde h_{2k} = d_{2k} + (X_k - \tilde X_k)$ tends to
$\eta + 2M(b - a)/(1 - ab)$ as $k \to \infty$, some $k$ gives
$h_{2k} - \tilde h_{2k} > \eta + 2M(b - a)/(1 - ab) - \varepsilon/2$. At $t = 2k$ we then have
$|h_t - \tilde h_t| > \eta + 2M\tanh(\rho/2) - \varepsilon$.
\end{proof}

\subsection{The log-timescale slopes}

\begin{lemma}[Log-timescale slopes]
\label{lem:param}
(i) For $\Delta = \mathrm{softplus}(z)$, the slope $\frac{d}{dz}\log\Delta$, which equals
$\operatorname{sigmoid}(z)/\mathrm{softplus}(z)$, lies in $(0, 1)$, is strictly decreasing in $z$
and tends to $1$ as $z \to -\infty$. Hence $|\log\Delta(z) - \log\Delta(\tilde z)| \le
|z - \tilde z|$, and for $z, \tilde z \ge z_{\mathrm{lo}}$ the constant $1$ may be replaced by
$\operatorname{sigmoid}(z_{\mathrm{lo}})/\mathrm{softplus}(z_{\mathrm{lo}})$.
(ii) For $\Delta = \Delta_{\min}(\Delta_{\max}/\Delta_{\min})^{\operatorname{sigmoid}(z)}$, the
slope $\frac{d}{dz}\log\Delta = \operatorname{sigmoid}'(z)\log(\Delta_{\max}/\Delta_{\min})$ lies
in $(0, \tfrac{1}{4}\log(\Delta_{\max}/\Delta_{\min})]$, with the maximum at $z = 0$.
\end{lemma}

\begin{proof}
For (i) we put $y = e^{z}$, for which $\operatorname{sigmoid}(z) = y/(1 + y)$ and
$\mathrm{softplus}(z) = \log(1 + y)$. Since $\log(1 + y) - y/(1 + y)$ vanishes at $y = 0$ and has
derivative $y/(1 + y)^2 > 0$, the ratio lies in $(0, 1)$. Its derivative has the sign of
$(1 - \operatorname{sigmoid})\,\mathrm{softplus} - \operatorname{sigmoid} =
(\log(1 + y) - y)/(1 + y) < 0$, and as $z \to -\infty$ both functions equal $e^{z}(1 + o(1))$.

For (ii), $\log\Delta = \log\Delta_{\min} + \operatorname{sigmoid}(z)\log(\Delta_{\max}/\Delta_{\min})$
and $\operatorname{sigmoid}' = \operatorname{sigmoid}(1 - \operatorname{sigmoid})$, which lies in
$(0, 1/4]$.
\end{proof}

Our float32 implementation clamps $\Delta$ to $[\Delta_{\min}, \Delta_{\max}]$ on the bounded
arm. In the reals the clamp is inactive, and it never increases $|\log\Delta - \log\tilde\Delta|$.

\subsection{The exponential-Euler cell}

With $x_t = \lambda\Delta_t$, $b_t = B_t u_t/\lambda$ and $\Psi$ as in Section~\ref{sec:euler} we
have $x_t = (1 - \Abar_t)\Psi(x_t)$, and the exponential-Euler
update $h_t = \Abar_t h_{t-1} + \Delta_t B_t u_t$ is the convex combination
\[
  h_t = \Abar_t h_{t-1} + (1 - \Abar_t)\, c^{\mathrm{ee}}_t,
  \qquad c^{\mathrm{ee}}_t = \Psi(x_t)\, b_t .
\]

\begin{corollary}[Exponential-Euler cell]
\label{cor:euler}
Suppose that $\Delta_t, \tilde\Delta_t \in (0, \Delta_{\mathrm{hi}}]$,
$|\log\Delta_t - \log\tilde\Delta_t| \le \rho$, $|\tilde b_t| \le M$ and
$|b_t - \tilde b_t| \le \eta^{b}$ for every $t$, and that
$h_{-1} = \tilde h_{-1} = 0$. With $G = \Psi(\lambda\Delta_{\mathrm{hi}})$, the
exponential-Euler cell satisfies $|h_t - \tilde h_t| \le G\eta^{b} + 3GM(e^{\rho} - 1)$ for every
$t \ge 0$.
\end{corollary}

\begin{proof}
Since $\log\Psi(x) = \log x - \log(1 - e^{-x})$, we have $d\log\Psi(x)/d\log x =
1 - x/(e^{x} - 1)$, which lies in $(0, 1)$ for $x > 0$. Hence $\Psi$ is increasing and
$|\log\Psi(x) - \log\Psi(\tilde x)| \le |\log x - \log\tilde x|$. With
$x_t, \tilde x_t \le \lambda\Delta_{\mathrm{hi}}$ this gives
$|\tilde c^{\mathrm{ee}}_t| = \Psi(\tilde x_t)|\tilde b_t| \le GM$ and $\Psi(x_t) \le G$, and
Theorem~\ref{thm:cell} applies with $GM$ in place of $M$.

The same bound gives $|\Psi(x_t) - \Psi(\tilde x_t)| \le (e^{\rho} - 1)\Psi(x_t)$, hence
$|c^{\mathrm{ee}}_t - \tilde c^{\mathrm{ee}}_t| \le \Psi(x_t)\,|b_t - \tilde b_t|
+ |\Psi(x_t) - \Psi(\tilde x_t)|\,|\tilde b_t| \le G\eta^{b} + (e^{\rho} - 1)GM$. Adding the
time-warp term $2GM(e^{\rho} - 1)$ gives the claim.
\end{proof}

\subsection{The block and the data-free block certificate}
\label{app:block}

We certify a block that computes
\begin{align*}
  \hat\xi       &= \mathrm{LN}(\xi),
  \quad [u_0, g] = \mathrm{in\_proj}(\hat\xi),
  \quad u = \mathrm{SiLU}\!\left( \mathrm{DepthwiseCausalConv1d}(u_0) \right), \\
  [p, B, C]     &= \mathrm{x\_proj}(u),
  \qquad \Delta = \Delta_{\min}\left( \Delta_{\max} / \Delta_{\min} \right)^{\operatorname{sigmoid}(\mathrm{dt\_proj}(p))}, \\
  A             &= -\exp(A_{\log}),
  \qquad
  \Abar_t = \exp(\Delta_t A),
  \qquad
  \Bbar_t = A^{-1}\!\left( \exp(\Delta_t A) - I \right) B_t, \\
  h_t           &= \Abar_t\, h_{t-1} + \Bbar_t\, u_t, \quad h_{-1} = 0,
  \qquad
  y_t = \textstyle\sum_n C_{t,n}\, h_{t,:,n} + D\, u_t, \\
  \mathrm{out}  &= \mathrm{out\_proj}\!\left( y \odot \mathrm{SiLU}(g) \right) + \xi,
\end{align*}
with all products elementwise and broadcast over the state axis; on the softplus arm
$\Delta = \mathrm{softplus}(\mathrm{dt\_proj}(p))$. Both $A = -\exp(A_{\log}) < 0$ and
$\Delta > 0$ hold by construction, for every checkpoint of the architecture.

With $\mathrm{LN}$ the layer normalisation of Lemma~\ref{lem:ln}, of width $d$, gain $\gamma$ and
offset $\beta$, we write $\mathcal{Z} = \{ \beta + \gamma \odot r : \mathbf{1}^{\top} r = 0,\
\|r\|_2 \le \sqrt{d} \}$ for the set on which the boxes of Section~\ref{sec:setting} are
certified, which contains $\mathrm{LN}(\xi)$ for every $\xi$, and $|W|$ for the entrywise absolute
value of a matrix $W$. We write $\hat u_i$, $\hat B_n$, $\hat C_n$ and $\hat s_i$ for the suprema
of $|u_i|$, $|B_n|$, $|C_n|$ and $|\mathrm{SiLU}(g_i)|$ over the certified boxes, $M_{i,n}$ for
the invariant radius of cell $(i, n)$, and $\hat y_i = \sum_n \hat C_n M_{i,n} + |D_i|\hat u_i$.
We write $s^{u}_i$ and $s^{g}_i$ for the suprema of $|\mathrm{SiLU}'|$ over the boxes of the
convolution output and the gate pre-activation of channel $i$, and $\ell_i$ for the supremum of
$d\log\Delta_i/dz_i$ over the box of $z_i$ (Lemma~\ref{lem:param}). Each $\Lambda$ below is a
sup-norm Lipschitz constant of the named stage over its certified box ($\Lambda^{u}_i$, for
example, bounds the change of $u_i$ per unit change of the normalised input).

\begin{corollary}[Block]
\label{cor:block}
Let $\xi$ and $\tilde\xi$ be two input sequences of the block, both runs starting from
$h_{-1} = 0$, and let $\|\mathrm{LN}(\xi_t) - \mathrm{LN}(\tilde\xi_t)\|_\infty \le \delta$ for
every $t$. Let $w_{i,\cdot}$ be the convolution kernel of channel $i$, let $W^{B}$, $W^{C}$ and
$W^{p}$ be the blocks of $\mathrm{x\_proj}$ that produce $B$, $C$ and $p$, let
$W^{\mathrm{in}}$, $W^{\mathrm{dt}}$ and $W^{\mathrm{out}}$ be the weights of $\mathrm{in\_proj}$,
$\mathrm{dt\_proj}$ and $\mathrm{out\_proj}$, and define
\begin{align*}
  \Lambda^{u}_i &= s^{u}_i\, \|w_{i,\cdot}\|_1 \textstyle\sum_j |W^{\mathrm{in}}_{ij}|,
  &
  \Lambda^{g}_i &= \textstyle\sum_j |W^{\mathrm{in}}_{d_{\mathrm{inner}} + i,\, j}|,
  \\
  \Lambda^{B} &= |W^{B}|\, \Lambda^{u},
  &
  \Lambda^{C} &= |W^{C}|\, \Lambda^{u},
  \\
  \Lambda^{z} &= |W^{\mathrm{dt}} W^{p}|\, \Lambda^{u},
  &
  \rho_i &= \ell_i \Lambda^{z}_i\, \delta,
  \\
  \eta_{i,n} &= \bigl( \hat B_n \Lambda^{u}_i + \hat u_i \Lambda^{B}_n \bigr)\, \delta / \lambda_{i,n},
  &
  E_{i,n} &= \eta_{i,n} + 2M_{i,n}\bigl(e^{\rho_i} - 1\bigr),
  \\
  Y_i &= \textstyle\sum_n \bigl( \hat C_n E_{i,n} + M_{i,n} \Lambda^{C}_n \delta \bigr)
          + |D_i|\, \Lambda^{u}_i\, \delta,
  &
  O_i &= \hat y_i\, s^{g}_i \Lambda^{g}_i\, \delta + \hat s_i\, Y_i,
  \\
  \Gamma_k &= \textstyle\sum_i |W^{\mathrm{out}}_{k i}|\, O_i .
\end{align*}
Then for every $t \ge 0$, channel $i$, state $n$ and output coordinate $k$,
$|\log\Delta_{t,i} - \log\tilde\Delta_{t,i}| \le \rho_i$,
$|c_{t,i,n} - \tilde c_{t,i,n}| \le \eta_{i,n}$, $|h_{t,i,n} - \tilde h_{t,i,n}| \le E_{i,n}$,
$|y_{t,i} - \tilde y_{t,i}| \le Y_i$ and
$|\mathrm{out}_{t,k} - \widetilde{\mathrm{out}}_{t,k}| \le \Gamma_k + |\xi_{t,k} - \tilde\xi_{t,k}|$.
\end{corollary}

\begin{proof}
We follow the perturbation through the stages in order. Since both normalised sequences lie in
$\mathcal{Z}$, every intermediate quantity of both runs, and every segment between
corresponding values, then lies in its certified box. Through the input projection, $u^{0}_i$ moves by
at most $\sum_j |W^{\mathrm{in}}_{ij}|\,\delta$ and $g_i$ by at most $\Lambda^{g}_i\delta$. With zero
left padding in both runs, the convolution multiplies that bound by $\|w_{i,\cdot}\|_1$, and the
mean value theorem for $\mathrm{SiLU}$ gives $\Lambda^{u}_i\delta$.

Linearity of $\mathrm{x\_proj}$ and $\mathrm{dt\_proj}$ gives $\Lambda^{B}_n\delta$, $\Lambda^{C}_n\delta$
and $\Lambda^{z}_i\delta$, and Lemma~\ref{lem:param} gives $\rho_i$. Writing
$B_n u_i - \tilde B_n\tilde u_i = B_n(u_i - \tilde u_i) + \tilde u_i(B_n - \tilde B_n)$ gives
$\eta_{i,n}$, and Theorem~\ref{thm:cell} gives $E_{i,n}$. The same product rule applied to
$C_n h_{i,n}$ with $|\tilde h_{i,n}| \le M_{i,n}$ gives $Y_i$. The gate, the output projection
and the residual then give $O_i$, $\Gamma_k$ and the term $|\xi_{t,k} - \tilde\xi_{t,k}|$.
\end{proof}

As a function of $\delta$, $\Gamma_k/\delta$ increases through $e^{\rho_i} - 1$ and tends, as
$\delta \to 0$, to the value with $\rho_i$ in place of $e^{\rho_i} - 1$. Perturbations that leave
$\mathcal{Z}$ need the boxes widened to $\mathcal{Z} + [-\delta, \delta]^{d}$
(Section~\ref{sec:exp-setup} gives the effect on our model).

\section{Relation to Cao et al.\ in detail}
\label{app:cao}

\paragraph{A derivation from their lemmas.}
In the setting of Cao et al.\ \citep{cao2026control}, $A$ is Hurwitz with
$\|e^{\sigma A}\| \le M_A e^{-\alpha\sigma}$, one timescale per step is shared by the whole state,
and by their Lemma~3.1 the scan samples a continuous flow at the grid of accumulated times. We
follow their argument for one cell of~\eqref{eq:convex}, whose flow is
$\dot h = -\lambda h + w$ with the forcing $w_s = \lambda c_s$ on step $s$. We take both runs
from the same state with $|\tilde h_{-1}| \le M$. With $\tau_{-1} = \tilde\tau_{-1} = 0$,
$\tau_t = \sum_{s \le t}\Delta_s$ and $\tilde\tau_t = \sum_{s \le t}\tilde\Delta_s$ we have
$h_t = h(\tau_t)$ and $\tilde h_t = \tilde h(\tilde\tau_t)$, where $h$ and $\tilde h$ solve
$\dot h = -\lambda h + w$ and $\dot{\tilde h} = -\lambda\tilde h + \tilde w$ from the common
start, with forcing $w_s$ on $[\tau_{s-1}, \tau_s)$ and $\tilde w_s = \lambda\tilde c_s$ on
$[\tilde\tau_{s-1}, \tilde\tau_s)$.

We map each interval $[\tilde\tau_{s-1}, \tilde\tau_s)$ linearly onto $[\tau_{s-1}, \tau_s)$ and
let $\hat g(\sigma)$ be the value of $\tilde h$ at the corresponding time, which gives
$\hat g(\tau_t) = \tilde h_t$ and, on $[\tau_{s-1}, \tau_s)$,
\[
  \dot{\hat g} = r_s\bigl(\tilde w_s - \lambda\hat g\bigr)
    = -\lambda\hat g + \tilde w_s + (r_s - 1)\bigl(\tilde w_s - \lambda\hat g\bigr),
  \qquad r_s = \tilde\Delta_s/\Delta_s .
\]
With the two runs now on one grid, their forcings differ by
$\varpi_s = w_s - \tilde w_s - (r_s - 1)(\tilde w_s - \lambda\hat g)$. Since
$|w_s - \tilde w_s| \le \lambda\eta$, $|r_s - 1| \le e^{\rho} - 1$, $|\tilde w_s| \le \lambda M$ and
$|\hat g| \le M$, we have
$|\varpi_s| \le \lambda\eta + 2\lambda M(e^{\rho} - 1)$. The bound $|\hat g| \le M$ holds because
between grid points $\tilde h$ is a convex combination of $\tilde h_{s-1}$ and $\tilde c_s$, with
$|\tilde h_{s-1}| \le M$ by Proposition~\ref{prop:invariant}. The mismatch depends on $\hat g$
itself, but that is admissible because $\hat g$ takes the values of $\tilde h$.

By variation of constants (the argument of their Lemma~B.2, which applies to any bounded
measurable forcing), two solutions of $\dot h = -\lambda h + w$ from a common start differ at time
$\sigma$ by at most $(1 - e^{-\lambda\sigma})/\lambda$ times the supremum of the difference of
their forcings. Since $e^{-\lambda\tau_t} = P_t$, this gives
$|h_t - \tilde h_t| \le (1 - P_t)(\eta + 2M(e^{\rho} - 1))$ for every $t$, which is
Theorem~\ref{thm:cell} for a common initial state.

In the proof of their Theorem~3.3, Cao et al.\ state that a direct bound on the forcing mismatch is
unavailable because the two forcings are piecewise constant on different grids, but our change of
time variable aligns the grids. Its size, the largest $|\log r_s|$, has the form of the norm that
Billingsley \citep[Section~12]{billingsley1999convergence} assigns to a time change of $[0, 1]$ in
the Skorokhod topology, here for a map between intervals of different lengths.

\paragraph{Comparison regime by regime.}
Their time-warp argument assumes one timescale shared by the whole state, which holds for each
channel of the block separately. Treating a diagonal $A$ one cell at a time is the most
favourable form of their bound, since their statement for the whole state uses
$\alpha = \min\lambda$ and $\|A\| = \max\lambda$. On the bounded arm, for example, the decay rates lie
between $0.912$ and $17.14$, while a statement for the whole state replaces all of them by the two
extremes. Both the additive baseline~\eqref{eq:additive} and
Theorem~\ref{thm:cell} pass through the readout in the same way, which lets us compare them at the
level of the state.

Without a floor, the additive time-warp term grows linearly in the number of steps and ours does
not, although for short sequences ours can be the larger. Let $n^{\ast}$ be the smallest
$n \ge 1$ at which the additive time-warp term without a floor after $n$ steps,
$2M\lambda\eta_\Delta n$, is at least our finite-horizon term at its worst over admissible
timescales, $2M(e^{\rho} - 1)(1 - e^{-\lambda\Delta_{\mathrm{hi}} n})$, to first order in the
size of the perturbation, where $\Delta_{\mathrm{hi}}$ is the certified ceiling. For fewer than
$n^{\ast}$ steps the additive term is the smaller. Because the finite-horizon term is at most
$2M(e^{\rho} - 1)$, $n^{\ast}$ is never later than the step at which the additive term passes the
flat bound of Theorem~\ref{thm:cell} in the same cell. Over the cells of the bounded arm
$n^{\ast}$ ranges from $3$ to $42$ with median $5$, while on the softplus arm it is at most $2$.

With a floor and only the additive information $\eta_\Delta$, the admissible value
$\rho = \log(1 + \eta_\Delta/\Delta_{\mathrm{lo}})$ turns our time-warp term into
$2\lambda M\eta_\Delta/x$ with $x = \lambda\Delta_{\mathrm{lo}}$. This lies between the additive
baseline $2\lambda M\eta_\Delta/(e^{x} - 1)$ and the form $2\lambda M\eta_\Delta/(1 - e^{-x})$ of
their Remark~3.4.
Under their hypotheses our term is thus no larger than theirs, and the ratios of the three terms
tend to $1$ as $x \to 0$. At
the certified floors of our trained model, where $x$ is below $0.018$, they differ by less than $2$
per cent.

We obtain the ratio of Remark~\ref{rem:cap} from the two terms directly. If $z$ changes by at most
$\zeta$ on the certified box $Z$ of its channel, then $\eta_\Delta = \zeta\sup_Z|\Delta'|$ and
$\rho = \zeta\sup_Z|(\log\Delta)'|$ are admissible. The limit $2M\lambda\eta_\Delta/(e^{x} - 1)$ of
the additive time-warp term, divided by our $2M(e^{\rho} - 1) = 2M\rho + O(\zeta^2)$, gives the
ratio of the remark. Its numerator $\sup_Z|\Delta'|$ is the counterpart, restricted to the box,
of the factor that multiplies $\|W_\Delta\|$ in the Lipschitz constant of $\Delta$ that Cao et al.\
assume (their Assumption~3.2(A4), bounded in their Proposition~4.6 and Remark~4.7 with the
supremum over all $z$). When $Z$ excludes $0$ the argument of the remark does not apply. Since $\Delta$ increases in $z$, on $Z$ we have
$|\Delta'| = \Delta\,|(\log\Delta)'| \le \Delta(\sup Z)\,|(\log\Delta)'|$, so with the floor
$\Delta_{\mathrm{lo}} = \Delta(\inf Z)$ that the box itself certifies, the first factor of the
ratio is at most $\Delta(\sup Z)/\Delta(\inf Z)$, a bound that tends to $1$ as the box narrows. For
the bounded map with $R = 100$ and $Z = [3, 4]$ the first factor itself equals $1$, because
$\sup_Z|\Delta'|$ and $\sup_Z|(\log\Delta)'|$ are both attained at $z = 3$ ($\Delta'$ peaks at
$z \approx 1.57$). The whole ratio is then
$x/(e^{x} - 1) < 1$, and the floored baseline is the tighter bound in such a cell.

At the empirical floor of Section~\ref{sec:floor-exp} (the smallest timescale in a channel that an
attack on the block's input reaches) the baseline is $8.5$ to $12$ times our bound at the output of
a bounded block. Per cell, however, the ratio there ranges from $0.16$ to $25$, and in about a
tenth of the cells an exact per-channel floor could make the additive term smaller than ours.

For the softplus arm the leading factor is about $1/\Delta_{\mathrm{lo}}$ when the floor is small.
With the certified floors of the trained softplus arm the per-cell ratio, which compares the limits
as $L \to \infty$, lies between $1.3 \times 10^{4}$ and $7.9 \times 10^{20}$, with medians of $9.1
\times 10^{10}$ and $4.0 \times 10^{10}$ on the two blocks.

\section{Interval arithmetic and the normalisation bounds}
\label{app:separate}

\subsection{The separated bound}

A verifier handed the update without~\eqref{eq:zoh} bounds $\Abar$ and $\Bbar$ separately and
obtains the geometric bound $\sup|\Bbar u|/(1 - \sup\Abar)$. In the true update the input term
and the contraction cancel, because the same factor $1 - \Abar_t$ multiplies the one and defines
the other. A
separated bound instead takes the input term at the ceiling $\Delta_{\mathrm{hi}}$ of the
timescale box and the contraction at its floor $\Delta_{\mathrm{lo}}$, which gives the per-cell
ratio to the invariant
\begin{equation}
\label{eq:gap}
  \mathrm{gap}(\lambda)
    = \frac{1 - e^{-\lambda \Delta_{\mathrm{hi}}}}{1 - e^{-\lambda \Delta_{\mathrm{lo}}}}.
\end{equation}
This ratio is at least $1$, with equality only for a one-point box. As
$\Delta_{\mathrm{lo}} \to 0$ it grows like $(1 - e^{-\lambda\Delta_{\mathrm{hi}}})/(\lambda\Delta_{\mathrm{lo}})$
while the invariant does not move, and since $(1 - e^{-y})/y$ is decreasing it never exceeds
$\Delta_{\mathrm{hi}}/\Delta_{\mathrm{lo}}$. It decreases in $\lambda$, because $y \mapsto y/(e^{y} - 1)$ is decreasing and the
log-derivative of~\eqref{eq:gap} in $\lambda$ is
$\Delta_{\mathrm{hi}}/(e^{\lambda\Delta_{\mathrm{hi}}} - 1) -
\Delta_{\mathrm{lo}}/(e^{\lambda\Delta_{\mathrm{lo}}} - 1) < 0$.

Interval arithmetic meets the same loss when it checks the induction step of
Proposition~\ref{prop:invariant}. Because the variable $a$ occurs twice in $a h + (1-a) c$, an
interval evaluator given a candidate box $M_h$ on the state and the bound $M_c$ on $c_t$ returns
$\sup\Abar\, M_h + (1 - \inf\Abar)\, M_c$. This check passes exactly when
$M_h \ge (1 - \inf\Abar)\, M_c / (1 - \sup\Abar)$, that is, for boxes at least the geometric
fixed point. For the invariant's own box, $M_h = M_c > 0$, it fails whenever the timescale box is
not a single point.

With $a$ kept as one variable, the supremum over $a \in [0,1]$ of $aM_h + (1-a)M_c$ is
$\max(M_h, M_c)$. The box is then preserved exactly when $M_c \le M_h$ (our implementation
checks the residual $\max(M_h, M_c) - M_h$).

On the bounded arm of our order book model, generic interval propagation iterated to convergence
agrees with the geometric fixed point, which is $95.6$ and $95.2$ times the invariant radius on
the two blocks.

\subsection{The normalisation bounds}

\begin{lemma}[Coordinate bound]
\label{lem:ln}
Let $\mathrm{LN}(x) = \gamma \odot (x - \mu)/\sqrt{\sigma^2 + \varepsilon} + \beta$ on
$\mathbb{R}^{d}$, with $d \ge 2$, $\varepsilon > 0$ and $\sigma^2$ the variance with divisor $d$.
Then $|\mathrm{LN}(x)_i - \beta_i| \le |\gamma_i|\sqrt{d - 1}$ for every $x$ and every $i$.
\end{lemma}

\begin{proof}
With $z = (x - \mu)/\sqrt{\sigma^2 + \varepsilon}$ we have $\sum_j z_j = 0$ and
$\sum_j z_j^2 = d\sigma^2/(\sigma^2 + \varepsilon) < d$. Cauchy--Schwarz on
$\sum_{j \ne i} z_j = -z_i$ gives $\sum_{j \ne i} z_j^2 \ge z_i^2/(d-1)$, hence
$d > z_i^2\, d/(d - 1)$ and $z_i^2 < d - 1$.
\end{proof}

With $\varepsilon = 0$, Lemma~\ref{lem:ln} is the Laguerre--Samuelson inequality
\citep{laguerre1880methode,samuelson1968deviant}, which bounds one coordinate at a time, whereas
the fused bound of Proposition~\ref{prop:fused} combines the zero-mean and norm constraints with
the following linear map and equals the supremum over all inputs.

\begin{proposition}[Fused bound]
\label{prop:fused}
Let $W_k$ be row $k$ of the linear map applied to the normalised output, let
$v_k = W_k \odot \gamma$ with $\bar{v}_k$ the mean of its entries, and let
$S = \{\, z : \mathbf{1}^{\top} z = 0,\ \|z\|_2 \le \sqrt{d} \,\}$. Then
$\sup_{z \in S} \langle v_k, z \rangle = \| v_k - \bar{v}_k \mathbf{1} \|_2 \sqrt{d}$, attained,
when $v_k$ is not constant, at
$z^{\ast} = \sqrt{d}\,(v_k - \bar{v}_k \mathbf{1}) / \|v_k - \bar{v}_k \mathbf{1}\|_2$. For
$\varepsilon > 0$ the supremum over the normalised vectors of all inputs is the same value, but
when $v_k$ is not constant it is not attained.
\end{proposition}

\begin{proof}
Since $\mathbf{1}^{\top} z = 0$, Cauchy--Schwarz over $\|z\|_2 \le \sqrt{d}$ applied to
$\langle v_k, z \rangle = \langle v_k - \bar{v}_k \mathbf{1},\, z \rangle$ gives the bound, which
$z^{\ast} \in S$ attains. Every normalised vector has
$\|z\|_2^2 = d\sigma^2/(\sigma^2 + \varepsilon) < d$, and when $v_k$ is not constant
Cauchy--Schwarz then gives
$\langle v_k, z \rangle \le \|v_k - \bar v_k\mathbf{1}\|_2\, \|z\|_2
< \|v_k - \bar v_k\mathbf{1}\|_2 \sqrt{d}$ for every normalised vector $z$, and the value is not
attained. For the input $x = t z^{\ast}$ the normalised vector is
$t z^{\ast}/\sqrt{t^2 + \varepsilon}$, which tends to $z^{\ast}$ as $t \to \infty$ and shows that
the supremum is the same value.
\end{proof}

Riechers \citep{riechers2024layernorm} and Brody et al.\ \citep{brody2023layernorm} describe this
geometry of layer normalisation (a projection onto the hyperplane orthogonal to $\mathbf{1}$
followed by a scaling). Bounding the normalisation and the linear map separately gives
$\|v_k\|_1\sqrt{d-1}$ instead, an $\ell_1$ quantity where an $\ell_2$ one suffices. This
difference carries over directly to the radius $\sup|Bu|/\lambda$, which is why our certificates
use the fused bound.

\end{document}